\pdfoutput=1
\documentclass{article}
\PassOptionsToPackage{numbers,sort&compress}{natbib}
\usepackage[preprint]{neurips_2026}
\usepackage[utf8]{inputenc}
\usepackage[T1]{fontenc}
\usepackage{amsmath,amssymb}
\usepackage{graphicx}
\usepackage{booktabs}
\usepackage{tabularx}
\usepackage{makecell}
\usepackage{bbm}
\usepackage{multirow}
\usepackage{placeins}
\usepackage{xurl}
\usepackage[hidelinks]{hyperref}
\hypersetup{
  pdftitle={CARDEA: Auditable Reasoning Grounded in Spatial Evidence for End-to-End Coronary Angiography Interpretation},
  pdfauthor={Jia-Jen Lee, Shih-Yen Hou, Kee Koon Ng, Wei-Chun Wang, Shih-Sheng Chang}
}
\usepackage{setspace}

\title{CARDEA: Auditable Reasoning Grounded\\
in Spatial Evidence for End-to-End\\
Coronary Angiography Interpretation}

\author{%
  Jia-Jen Lee\textsuperscript{a},
  Shih-Yen Hou\textsuperscript{a},
  Kee Koon Ng\textsuperscript{b},
  Wei-Chun Wang\textsuperscript{a,c,d},
  Shih-Sheng Chang\textsuperscript{a,b,e} \\[0.5em]
  \normalfont
  \textsuperscript{a}Artificial Intelligence and Robotics Innovation Center, \\
  China Medical University Hospital, Taichung, Taiwan \\
  \textsuperscript{b}Division of Cardiovascular Medicine, Department of Internal Medicine, \\
  China Medical University Hospital, Taichung, Taiwan \\
  \textsuperscript{c}Department of Neurology, China Medical University Hospital, \\
  China Medical University, Taichung, Taiwan \\
  \textsuperscript{d}Neuroscience and Brain Disease Center, China Medical University, Taichung, Taiwan \\
  \textsuperscript{e}School of Medicine, China Medical University, Taichung, Taiwan}

\date{}

\begin{document}

\maketitle

\begin{abstract}
Invasive coronary angiography (CAG) is the gold standard for diagnosing coronary artery disease, but interpretation varies substantially among observers. Existing AI systems can improve consistency but lack auditable decision processes and are limited in comprehensive open-ended assessment, undermining clinician trust and clinical adoption readiness. We developed CARDEA, a unified large vision-language model that serves as the inference core of a CAG pipeline. It was trained solely on public datasets and closed-ended tasks in three stages: visual feature alignment, a self-distilled Chain-of-Box (CoB) cold start, and reinforcement learning with verifiable rewards (RLVR) with a CoB reward encouraging bounding-box use in the reasoning trace. We assessed its two study-level diagnoses, dominance classification and complexity assessment, against a dedicated classifier and two interventional cardiologists. Report generation was excluded from training and evaluated zero-shot across stages on an external cohort using vessel-severity macro-$F_1$. CARDEA trailed the classifier on in-distribution dominance but drew level under domain shift (accuracy, 0.91 [95\% confidence interval (CI), 0.86 to 0.95]) and was comparable to the cardiologists on complexity assessment (accuracy, 0.90 [CI, 0.82 to 0.97]). Only RLVR improved zero-shot report generation, raising its vessel-severity macro-$F_1$ (0.686 [CI, 0.664 to 0.707]) above the untuned base model (0.513) and over twice the always-normal floor (0.312). CARDEA runs an end-to-end CAG pipeline from raw multi-view videos through keyframe selection to study-level diagnosis while exposing auditable spatial evidence behind its conclusions. RLVR on verifiable closed-ended tasks surfaced open-ended reporting ability that supervised imitation did not. Clinical use requires prospective validation against expert cardiologists.
\end{abstract}

\textbf{Keywords:} Coronary Angiography, End-to-End Pipeline, Large Vision-Language Models, Reinforcement Learning with Verifiable Rewards, Auditable Reasoning, Visual Grounding.

\section{Introduction}
\label{sec:introduction}

Invasive coronary angiography (CAG) remains the gold standard for diagnosing coronary artery disease \cite{oikonomou2022current}. However, projecting 3D coronary structures onto a 2D plane introduces geometric distortions such as vessel overlap and foreshortening, which can lead to underestimation of lesion severity \cite{cimen2016reconstruction}. Because acquiring additional projections increases contrast exposure and radiation, clinicians must mentally integrate the available views of each lesion \cite{green2016optimal}. Visual interpretation remains subject to inter-observer variability, with a recent study reporting 77.4\% overall agreement among three experienced cardiologists reading the same angiograms \cite{shivaie2024interobserver}.

Deep learning has been applied to reduce this variability, with early single-view models analyzing individual angiographic projections \cite{chen2025computational, moon2021automatic}; cascaded pipelines like CathAI \cite{avram2023cathai} and DeepCoro \cite{labrecque2024evaluation} chained several such modules, but remain vulnerable to compounding errors across them. To address this, foundation models such as DeepCORO-CLIP \cite{harrabi2026deepcoroclip} pretrain on video-text pairs and fuse multiple views for study-level analysis. Yet these models are discriminative, outputting only closed-ended labels or spatial coordinates. Recent work including Nakamura et al.\ \cite{nakamura2025cag} and Jiang et al.\ \cite{jiang2026vision} has begun applying large vision-language models (LVLMs) \cite{li2023llava} to CAG for free-text diagnosis or reporting. However, such open-ended drafts are clinically useful only when easy to verify; pairing a diagnosis with spatial anchors lowers verification cost \cite{bannur2024maira} while raising clinician trust \cite{yildirim2024multimodal}. In practice, neither does this within the report. Nakamura et al.'s model cannot output bounding boxes, and Jiang et al.'s appear only in sub-tasks separate from the report. Existing systems thus leave a gap: they either remain discriminative or generate open-ended narratives without auditable traces that let clinicians inspect a model's logic.

Reinforcement learning with verifiable rewards (RLVR), exemplified by DeepSeek-R1 \cite{deepseek2025r1}, elicits reasoning traces that add interpretability. But an ordinary text-only trace does not reveal which anatomical structures the model attends to; grounded-reasoning work therefore embeds bounding boxes as spatial anchors within the reasoning trace \cite{fan2025grit}. A related study brings this to medical imaging and terms this mechanism Chain-of-Box (CoB) \cite{xu2025medground}, letting a human audit how each conclusion was reached.

We present CARDEA, a unified LVLM that runs an end-to-end CAG pipeline, from raw multi-view sequences through keyframe selection to study-level diagnosis, coupling competitive diagnostic performance with auditable CoB reasoning. Trained only on closed-ended tasks from public datasets \cite{jimenez2024cadica, popov2024dataset, kruzhilov2025coronarydominance, ponomarchuk2025cardiosyntax} through three stages---visual feature alignment, a self-distilled CoB cold start, and RLVR with a CoB reward---we evaluate whether this competence generalizes zero-shot to a fully held-out cohort, including open-ended report generation. We also examine whether policy optimization surfaces clinical reasoning that supervised imitation does not. The model weights and inference code are released for reproducibility.

\section{Methods}
\label{sec:methods}

\begin{figure*}[t]
\centering
\includegraphics[width=0.90\textwidth]{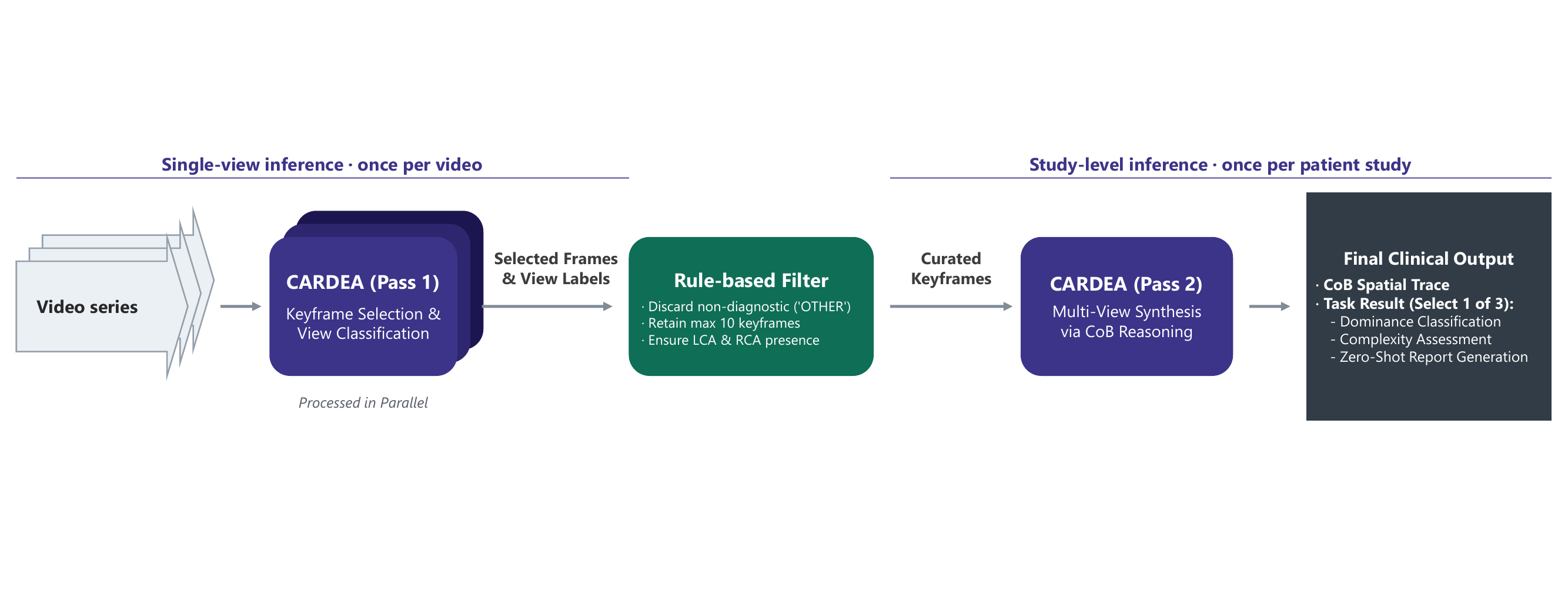}
\caption{End-to-end inference pipeline of CARDEA. A first pass filters non-diagnostic frames in parallel via keyframe selection and view classification; a second pass applies Chain-of-Box (CoB) reasoning over the curated multi-view keyframes to produce Dominance Classification, Complexity Assessment, and zero-shot Report Generation.}
\label{fig:pipeline}
\end{figure*}

CARDEA interprets a full CAG study through a two-pass pipeline (Figure~\ref{fig:pipeline}). A first pass runs single-view inference on each angiographic video to select representative keyframes and classify their view; a second pass performs study-level multi-image CoB reasoning over the curated keyframes. The first pass filters raw frames because processing them all with a large model is computationally prohibitive, and many are non-diagnostic owing to poor cardiac alignment or insufficient contrast. It therefore discards these frames and retains a compact set covering the major left and right coronary views.

\subsection{Datasets and Tasks}
To equip CARDEA to perform every task of the two-pass pipeline itself and to ground its reasoning in CoB evidence, we reformulated five public datasets into instructional tasks; full preprocessing and splits are in Appendix~\ref{sec:appendix_datasets}. To align the model with foundational CAG features and let it emit meaningful bounding boxes within its reasoning, ARCADE \cite{popov2024dataset} (3,000 keyframes from 1,500 patients with diverse equipment) supplies two single-view tasks, \textit{Vessel Detection} (25 coronary segments based on the SYNTAX score~\cite{sianos2005syntax}) and \textit{Stenosis Detection} ($\ge 50\%$ diameter stenosis). For the single-view first pass, which curates diagnostic frames, CADICA \cite{jimenez2024cadica} (multi-view videos from 42 patients) supplies \textit{Keyframe Selection}. The first pass then performs \textit{View Classification}, for which ARCADE's vessel annotations supply the left coronary artery (LCA) and right coronary artery (RCA) classes, while non-diagnostic CADICA frames and non-CAG medical images from PubMedVision \cite{chen2024huatuogpt} supply the OTHER class. For the second pass, which produces the study-level diagnoses, CoronaryDominance \cite{kruzhilov2025coronarydominance} (1,574 studies) supplies \textit{Dominance Classification} (Left vs.\ Right, by the SYNTAX definition), and CardioSyntax \cite{ponomarchuk2025cardiosyntax} (1,844 studies) supplies \textit{Complexity Assessment}, which we define by discretizing the continuous SYNTAX score into normal-to-intermediate ($0$--$32$) vs.\ high ($> 32$) complexity.

AngioCAD \cite{hosseini2026angiocad} is excluded from all training and serves as our held-out zero-shot benchmark. It supplies \textit{Multi-Frame View Classification} (LCA vs.\ RCA), \textit{Multi-Frame RCA Binary Stenosis Classification} (Lesion vs.\ Non-lesion), and open-ended \textit{Report Generation} over the four major branches: left main (LM), left anterior descending (LAD), left circumflex (LCX), and RCA.

\subsection{Model and Training}
\begin{figure*}[t]
\centering
\includegraphics[width=0.90\textwidth]{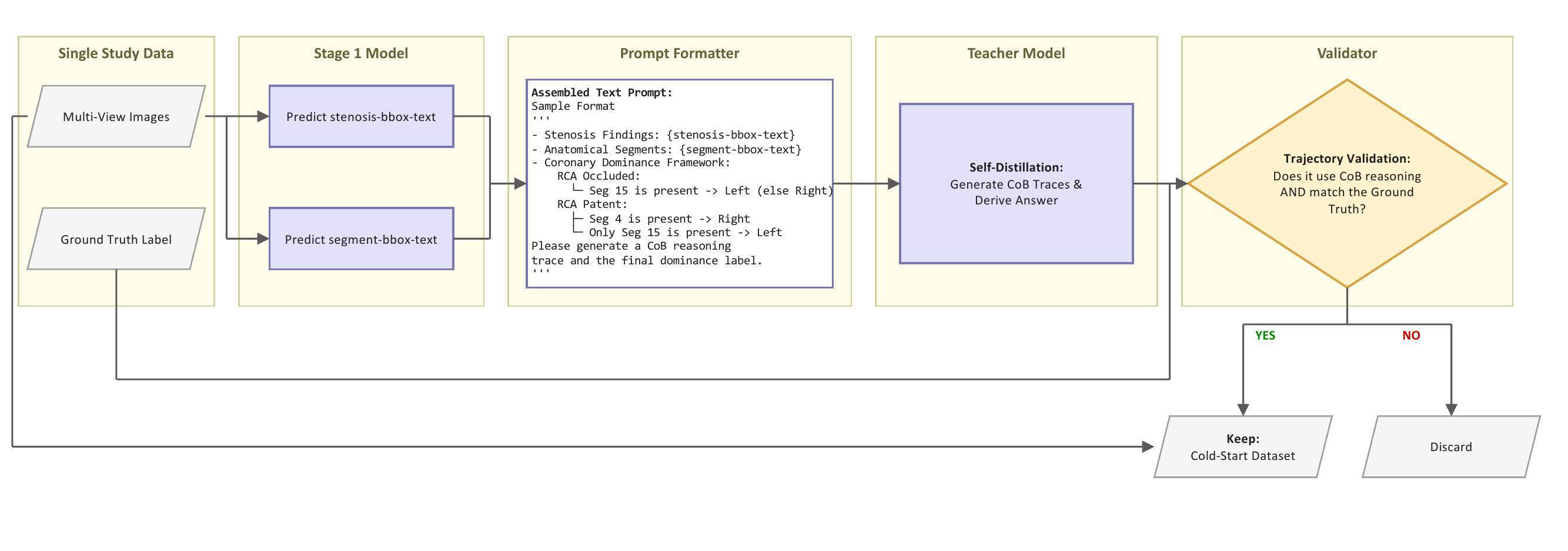}
\caption{Cold-start data preparation for Stage 2 training. The Stage 1 model generates vessel and stenosis bounding boxes for each keyframe; a teacher model synthesizes CoB reasoning traces conditioned on these detections; only traces reaching the correct diagnostic conclusion are retained for generative fine-tuning.}
\label{fig:coldstart_pipeline}
\end{figure*}

CARDEA is built on Qwen3-VL-30B-A3B-Thinking \cite{bai2025qwen3vltechnicalreport} and trained in three stages: two of supervised fine-tuning (SFT), followed by RLVR. Configuration and hyperparameters are in Appendix~\ref{sec:appendix_training}. The first stage aligns the model from the general domain to CAG. We fine-tune it on the single-view tasks, teaching it to localize vessels and stenoses with bounding boxes. The second stage builds on the first, extending the model's perception into its reasoning. However, CoB is not native to the base model, and hand-annotating reasoning traces is costly. We therefore synthesize the cold-start data by self-distillation \cite{chen2025metis} on dominance classification. The untuned base model serves as the teacher, generating the CoB reasoning traces from the Stage 1 model's boxes and a textual dominance decision guide (Figure~\ref{fig:coldstart_pipeline}). The third stage uses RLVR to push accuracy on study-level tasks and to strengthen the CoB behavior seeded by the cold start. Reward functions are defined in Appendix~\ref{sec:appendix_rewards}. For study-level tasks, CoB behavior earns an extra reward conditional on a correct final answer. We restrict it to the study level, because these diagnoses draw a single conclusion from several views and therefore need CoB to explain how they reach that conclusion from local features across the views. Conversely, a single-view task involves no cross-view synthesis and needs no such explanation. We deliberately exclude open-ended tasks like report generation from RLVR. Such tasks have no rule-based verifier, so they would need a reward model, which is prone to reward hacking \cite{deepseek2025r1}.

\subsection{Metrics and Baselines}
We choose evaluation metrics by task type. Classification uses accuracy and Macro $F_1$, with a target-class $F_1$ for the binary AngioCAD tasks to match with the baselines' report; detection uses instance-level $F_1$@IoU$\ge$0.5. Two tasks use custom metrics: keyframe selection by a mean frame distance ($D_k$), and report generation by Vessel Severity Macro-$F_1$ (VS-$F_1$) in two-class and three-class forms, where MedGemma-27B-IT \cite{sellergren2025medgemma} parses each free-text report into per-vessel severity labels. ROUGE-L~\cite{lin2004rouge} and BERTScore~\cite{zhang2019bertscore} serve as reference-based text-similarity metrics. Metric definitions are in Appendix~\ref{sec:appendix_eval_metrics}.

Baselines are published dedicated models evaluated on the same test sets \cite{pascual2025hyperparameter, labrecque2024evaluation, tran2025anatomy, kruzhilov2025coronarydominance, hosseini2026angiocad}, named per task in Tables~\ref{tab:foundational_tasks} and~\ref{tab:angiocad_closed}, with details in their footnotes. The exception is complexity assessment, where we compare against two interventional cardiologists with 10 and 3 years of experience \cite{ponomarchuk2025cardiosyntax}. Keyframe selection has no published baseline, so we report CARDEA's value alone.

Some tasks are reported under several settings. For stenosis detection, we additionally report $F_1$@(IoU$\ge$0.5 or IoP$\ge$0.6), the relaxed matching criterion of Jiang et al.'s LVLM~\cite{jiang2026vision}, which reduces sensitivity to differences in box extent and makes our value comparable with theirs (Appendix~\ref{sec:appendix_f1iou}). For vessel detection, we also report an 11-segment result matching DeepCoro's original convention. For view classification, CARDEA trains on three classes (including OTHER), but the baseline was evaluated only on LCA and RCA frames, so we add a two-class result to match. For dominance classification, we report on two official subsets: in-distribution Real Distribution (clinical class imbalance) and out-of-distribution Domain Shift (distinct imaging equipment). On held-out AngioCAD, to gauge the pipeline's view filtering, we report the full cohort ($n{=}412$) and a valid-views subset ($n{=}326$) retaining studies with both LCA and RCA views identified by CARDEA; the subset is for reference only because the baselines were not evaluated on it.

For every diagnostic metric we report a 95\% bootstrap confidence interval (CI), whereas for text-similarity metrics we report point estimates only. Two estimates are distinguishable when their CIs do not overlap, or a bare baseline estimate falls outside CARDEA's CI, and otherwise comparable, though not necessarily equivalent. Because these CIs are not adjusted for multiple comparisons, all distinctions are exploratory. Two further caveats apply. First, two test sets are small (complexity, $n{=}60$; keyframe selection, 48 videos from 5 patients) and may be underpowered. Second, two comparisons are not strictly head-to-head: DeepCoro's Algorithm~4 outputs segmentation masks that we converted to bounding boxes, possibly understating its detection score, and the cardiologists' binary labels come from a post-hoc binarization of continuous SYNTAX scores, not a task they performed natively.

\section{Results}
\label{sec:results}

\subsection{Closed-Ended Diagnostics}
\label{sec:foundational_perception}

Throughout, CARDEA denotes our final model in thinking mode, which enables explicit CoB reasoning. On the tasks it was trained on (Table~\ref{tab:foundational_tasks}), CARDEA was competitive with specialized baselines. In single-view detection it reached 0.37 on stenosis (vs.\ 0.36 for DCA-YOLOv8) and 0.50 on vessel detection (vs.\ 0.47 for DeepCoro's Algorithm~4). It reached a Macro $F_1$ of 0.99 on both the 3-class and 2-class view tasks (the latter vs.\ 1.00 for YOLOv8x-cls), and a mean frame distance of 0.60 on keyframe selection, placing its predicted optimal frame within one frame of the expert-annotated usable range on average, though this split (48 videos, 5 patients) is preliminary. At the study level, dominance classification was lower than the 2D ConvNeXt on the Real Distribution subset (accuracy, 0.94 vs.\ 0.97; Macro $F_1$, 0.88 vs.\ 0.94) but comparable under Domain Shift (accuracy, 0.91 vs.\ 0.89; Macro $F_1$, 0.89 vs.\ 0.88). On complexity assessment CARDEA matched two cardiologists in accuracy (0.90 vs.\ 0.90 and 0.88), though its Macro $F_1$ point estimate was slightly lower (0.80 vs.\ 0.83 and 0.81). Across these closed-ended tasks, CARDEA's only statistically distinguishable shortfall was Real Distribution dominance.

In contrast to the task-specific baselines, we additionally compared CARDEA with Jiang et al.'s LVLM~\cite{jiang2026vision}. For stenosis detection, CARDEA's $F_1$@(IoU$\ge$0.5 or IoP$\ge$0.6) of 0.60 matched their 0.60, up from its $F_1$@IoU$\ge$0.5 of 0.37. For vessel detection, its $F_1$@IoU$\ge$0.5 of 0.50 exceeded their 0.46.

\begin{table*}[htbp]
\centering
\caption{Trained Closed-Ended Performance. Bold marks the better value in rows where the two are statistically distinguishable: the comparator's 95\% CI does not overlap CARDEA's 95\% bootstrap CI (a baseline reported without a CI is treated as a zero-width interval, so its point estimate must fall outside CARDEA's CI). These intervals are not adjusted for multiple comparisons; the resulting distinctions should therefore be read as exploratory rather than as confirmatory hypothesis tests.}
\label{tab:foundational_tasks}
\small
\renewcommand{\arraystretch}{1.25}
\setlength{\tabcolsep}{5pt}
\begin{tabular}{@{}ll l c l@{}}
\toprule
\textbf{Level} & \textbf{Task} & \textbf{Metric} & \textbf{CARDEA} & \textbf{Baseline / Expert} \\
\midrule
\multirow{10}{*}{\textit{Frame}}
 & \multirow{2}{*}{Stenosis Det.}
   & $F_1$@IoU$\ge$0.5           & 0.37 (0.32--0.41) & $0.36 \pm 0.08$~\textsuperscript{a} \\
 & & $F_1$@(IoU$\ge$0.5 or IoP$\ge$0.6)
                                 & 0.60 (0.56--0.65) & 0.60~\textsuperscript{f} \\
\addlinespace[0.4ex]
 & \multirow{3}{*}{Vessel Det.}
   & $F_1$@IoU$\ge$0.5, 25 seg.  & 0.50 (0.48--0.53) & 0.47 (0.45--0.50)~\textsuperscript{b,\,$\dagger$} \\
 & & $F_1$@IoU$\ge$0.5, 25 seg.  & \textbf{0.50} (0.48--0.53) & 0.46~\textsuperscript{f} \\
 & & $F_1$@IoU$\ge$0.5, 11 seg.  & 0.58 (0.54--0.61) & 0.57 (0.54--0.61)~\textsuperscript{b,\,$\dagger$} \\
\addlinespace[0.4ex]
 & \multirow{2}{*}{View Cls.\ (3-class)}
   & Accuracy                    & 0.99 (0.98--1.00) & --- \\
 & & Macro $F_1$                 & 0.99 (0.98--1.00) & --- \\
\addlinespace[0.4ex]
 & \multirow{2}{*}{View Cls.\ (2-class)}
   & Accuracy                    & 0.99 (0.98--1.00) & 1.00~\textsuperscript{c} \\
 & & Macro $F_1$                 & 0.99 (0.98--1.00) & 1.00~\textsuperscript{c} \\
\addlinespace[0.4ex]
 & Keyframe Sel.
   & $D_k$~($\downarrow$)        & 0.60 (0.23--1.08) & --- \\
\cmidrule(l){1-5}
\multirow{6}{*}{\textit{Study}}
 & \multirow{2}{*}{Dominance (Real)}
   & Accuracy                    & 0.94 (0.92--0.96) & \textbf{0.97}~\textsuperscript{d} \\
 & & Macro $F_1$                 & 0.88 (0.82--0.92) & \textbf{0.94}~\textsuperscript{d} \\
\addlinespace[0.4ex]
 & \multirow{2}{*}{Dominance (Shift)}
   & Accuracy                    & 0.91 (0.86--0.95) & 0.89~\textsuperscript{d} \\
 & & Macro $F_1$                 & 0.89 (0.83--0.94) & 0.88~\textsuperscript{d} \\
\addlinespace[0.4ex]
 & \multirow{2}{*}{Complexity Assess.}
   & Accuracy                    & 0.90 (0.82--0.97)
   & \makecell[tl]{E1: 0.90 (0.82--0.97)~\textsuperscript{e,\,$\ddagger$}\\ E2: 0.88 (0.80--0.95)} \\
 & & Macro $F_1$                 & 0.80 (0.63--0.93)
   & \makecell[tl]{E1: 0.83 (0.68--0.94)~\textsuperscript{e,\,$\ddagger$}\\ E2: 0.81 (0.66--0.92)} \\
\bottomrule
\end{tabular}

\vspace{0.5ex}
{\footnotesize\setstretch{1}\raggedright
Entries are point estimates with 95\% confidence intervals in parentheses, obtained by recomputing each metric across 5{,}000 case-level bootstrap resamples. A baseline is shown as a point estimate alone where the source reported no interval and its per-case predictions were unavailable, so no interval could be computed here; the one exception is DCA-YOLOv8, whose source-reported uncertainty is given as $\pm$ (footnote a).\par
Macro $F_1$, unweighted mean of per-class $F_1$; $F_1$@IoU$\ge$0.5, instance-level detection $F_1$ at an IoU threshold of 0.5; IoP, intersection over prediction; 25 seg.\ and 11 seg., the 25 ARCADE-defined coronary segments and DeepCoro's original 11-segment convention; $D_k$, mean distance (in frames) from the predicted keyframe to the expert-annotated usable range, lower being better. OTHER, non-diagnostic or non-CAG frames; View Cls.\ (3-class) covers LCA/RCA/OTHER and the 2-class variant LCA/RCA only. Real and Shift are the official in-distribution (clinical class imbalance) and out-of-distribution (distinct imaging equipment) CoronaryDominance test subsets.\par
\textsuperscript{a} DCA-YOLOv8, tuned by CMA-ES, the best optimizer on the large backbone~\cite{pascual2025hyperparameter}. The uncertainty is the source-reported 95\% CI from stratified 3-fold cross-validation, not a case-level bootstrap CI.\par
\textsuperscript{b} DeepCoro's Algorithm~4, an ensemble of seven vessel-segmentation models~\cite{labrecque2024evaluation}.\par
\textsuperscript{c} YOLOv8x-cls, a YOLO model specifically trained for view classification~\cite{tran2025anatomy}.\par
\textsuperscript{d} 2D ConvNeXt, classifying frames independently with majority-vote aggregation~\cite{kruzhilov2025coronarydominance}.\par
\textsuperscript{e} E1 / E2 = Expert 1 (10 yrs) / Expert 2 (3 yrs), interventional cardiologists~\cite{ponomarchuk2025cardiosyntax}.\par
\textsuperscript{f} Jiang et al.'s large vision-language model~\cite{jiang2026vision}. Their relaxed criterion additionally accepts a predicted box whose intersection over prediction with an unmatched ground-truth box is $\ge 0.6$ (Appendix~\ref{sec:appendix_f1iou}); following their rationale, we apply it to stenosis only, where the location of a box matters more than its exact extent. Their vessel score uses the standard $F_1$@IoU$\ge$0.5.\par
\textsuperscript{$\dagger$} Algorithm~4's masks were converted to bounding boxes for this detection score, possibly understating it. In both the 25- and 11-segment comparisons, CARDEA and DeepCoro were evaluated over the same segments.\par
\textsuperscript{$\ddagger$} Derived by binarizing the experts' continuous SYNTAX scores at the $\le 32$ vs.\ $> 32$ threshold.\par}
\end{table*}

\subsection{Zero-Shot Generalization on AngioCAD}
\label{sec:zero_shot_generalization}

On the closed-ended AngioCAD tasks (Table~\ref{tab:angiocad_closed}), CARDEA was evaluated zero-shot against two baselines developed on this held-out cohort \cite{hosseini2026angiocad}. It surpassed the Adaptive Feature Fusion baseline on RCA binary stenosis classification (Lesion $F_1$, 0.85 vs.\ 0.81) and trailed the VGG19+LSTM model on view classification (RCA $F_1$, 0.90 vs.\ 0.95). Restricting to the valid-views subset raised both scores (Lesion $F_1$, 0.86; RCA $F_1$, 0.95); as this quality filter is not applied to the baselines, these results are shown for reference only.

\begin{table*}[htbp]
\centering
\caption{Zero-Shot Closed-Ended Performance on the External AngioCAD Dataset. Bold marks the better value in rows where the baseline's point estimate lies outside CARDEA's 95\% bootstrap CI. These intervals are not adjusted for multiple comparisons; the resulting distinctions should therefore be read as exploratory rather than as confirmatory hypothesis tests.}
\label{tab:angiocad_closed}
\small
\begin{tabular}{lccc}
\toprule
\textbf{Task \& Model} & \textbf{Accuracy} & \textbf{Target $F_1$} & \textbf{Macro $F_1$} \\
\midrule
\multicolumn{4}{l}{\textit{RCA Stenosis Classification --- Full}} \\
Adaptive Feature Fusion\textsuperscript{a} & 0.72 & 0.81\textsuperscript{$\dagger$} & --- \\
CARDEA & \textbf{0.80} (0.77--0.82) & \textbf{0.85} (0.83--0.87)\textsuperscript{$\dagger$} & 0.77 (0.73--0.80) \\
\midrule
\multicolumn{4}{l}{\textit{RCA Stenosis Classification --- Valid-views}} \\
CARDEA & 0.81 (0.78--0.84) & 0.86 (0.84--0.88)\textsuperscript{$\dagger$} & 0.78 (0.75--0.81) \\
\midrule
\multicolumn{4}{l}{\textit{View Classification --- Full}} \\
VGG19 + LSTM\textsuperscript{b} & \textbf{0.97} & \textbf{0.95}\textsuperscript{$\ddagger$} & --- \\
CARDEA & 0.94 (0.93--0.95) & 0.90 (0.88--0.92)\textsuperscript{$\ddagger$} & 0.93 (0.92--0.94) \\
\midrule
\multicolumn{4}{l}{\textit{View Classification --- Valid-views}} \\
CARDEA & 0.97 (0.96--0.98) & 0.95 (0.94--0.96)\textsuperscript{$\ddagger$} & 0.96 (0.96--0.97) \\
\bottomrule
\end{tabular}

\vspace{0.5ex}
{\footnotesize\setstretch{1}\raggedright
Entries are point estimates with 95\% confidence intervals in parentheses, obtained by recomputing each metric across 5{,}000 case-level bootstrap resamples. Neither baseline reported an interval and their per-case predictions were unavailable, so both are shown as point estimates alone.\par
Both tasks operate on 5-frame clips and are binary: RCA stenosis classification (Lesion vs.\ Non-lesion) and view classification (LCA vs.\ RCA). Full comprises all 412 studies and excludes none, so both models are scored on the same cohort; the valid-views subset ($n{=}326$) keeps only studies in which CARDEA identified both an LCA and an RCA view, a filter the baselines do not apply, and is therefore shown for reference only (Section~\ref{sec:zero_shot_generalization}). Each clip is anchored on a keyframe CARDEA selects itself, whereas the baselines follow their original frame-sampling protocol, a difference intrinsic to comparing an end-to-end system against modular baselines.\par
\textsuperscript{a} Dynamic weighting of dual-backbone features (ResNet101 and VGG16)~\cite{hosseini2026angiocad}.\par
\textsuperscript{b} Frame-wise feature extraction (VGG19) with temporal sequence modeling (LSTM)~\cite{hosseini2026angiocad}.\par
\textsuperscript{$\dagger$} Target metric represents the lesion-positive class (Lesion) $F_1$-score.\par
\textsuperscript{$\ddagger$} Target metric represents the minority class (RCA) $F_1$-score.\par}
\end{table*}

\subsection{Report Generation Across Training Stages}
\label{sec:emergence}

A core question is whether closed-ended training alone can improve an LVLM's open-ended report generation; in our pipeline this ability rose only after RLVR, not under SFT (Table~\ref{tab:stage_wise_evolution}). Since text-similarity metrics need not solely reflect diagnostic agreement, we prioritize VS-$F_1$. An always-normal report in the reference format served as a stress test for text-similarity metrics and as the VS-$F_1$ floor. With 74\% of graded AngioCAD sub-segments normal, it outscored our final model on ROUGE-L (0.81 vs.\ 0.35) and BERTScore (0.77 vs.\ 0.74), despite a two-class VS-$F_1$ of 0.312. Because VS-$F_1$ scores extracted discrete severity labels and macro-averages across classes, wording overlap and normal-class dominance affect it less. The untuned base model already reached a two-class VS-$F_1$ of 0.513, well above the naive floor, but the supervised stages eroded it (Stage 1, 0.452; Stage 2, 0.373). Only RLVR reversed the decline, raising it to 0.644 with thinking disabled and 0.686 with native thinking, surpassing the base model and more than doubling the always-normal floor; the three-class score mirrored this trajectory. On the valid-views subset the final two-class VS-$F_1$ reached 0.716.

\begin{table*}[htbp]
\centering
\caption{Stage-Wise Evolution of Zero-Shot Report Generation Quality on AngioCAD. Bold marks CARDEA (Final) on the full cohort, the result reported throughout the paper.}
\label{tab:stage_wise_evolution}
\small
\setlength{\tabcolsep}{3.5pt}
\begin{tabular}{l cc cc}
\toprule
 & \multicolumn{2}{c}{\textbf{Diagnostic}} & \multicolumn{2}{c}{\textbf{Text similarity}} \\
\cmidrule(lr){2-3}\cmidrule(lr){4-5}
\textbf{Stage} & \textbf{VS-$F_1$ (2-class)} & \textbf{VS-$F_1$ (3-class)} & \textbf{ROUGE-L} & \textbf{BERTScore} \\
\midrule
\multicolumn{5}{l}{\textit{Full cohort ($n{=}412$)}} \\
Naive (always-normal)    & 0.312 (0.298--0.325) & 0.208 (0.199--0.217) & 0.81 & 0.77 \\
Base Model (Qwen3-VL)~\cite{bai2025qwen3vltechnicalreport} & 0.513 (0.481--0.544) & 0.361 (0.335--0.389) & 0.22 & 0.66 \\
Stage 1 (Align)          & 0.452 (0.427--0.478) & 0.309 (0.287--0.332) & 0.33 & 0.73 \\
Stage 2 (Cold start)     & 0.373 (0.351--0.396) & 0.253 (0.235--0.274) & 0.34 & 0.69 \\
Stage 3 (RLVR w/o Think) & 0.644 (0.619--0.668) & 0.480 (0.454--0.505) & 0.35 & 0.73 \\
\textbf{CARDEA (Final)}  & \textbf{0.686} (0.664--0.707) & \textbf{0.508} (0.485--0.531) & 0.35 & 0.74 \\
\midrule
\multicolumn{5}{l}{\textit{Valid-views subset ($n{=}326$)}} \\
CARDEA (Final)           & 0.716 (0.692--0.739) & 0.549 (0.522--0.575) & 0.36 & 0.74 \\
\bottomrule
\end{tabular}

\vspace{0.5ex}
{\footnotesize\setstretch{1}\raggedright
Entries in the two VS-$F_1$ columns are point estimates with 95\% confidence intervals in parentheses, obtained by recomputing each metric across 5{,}000 case-level bootstrap resamples. The text-similarity columns are point estimates only.\par
All rows except the last use the full cohort of 412 studies, which excludes none; the last row reports CARDEA on the valid-views subset ($n{=}326$), which keeps only studies in which both an LCA and an RCA view were identified.\par
Always-normal is a fixed dummy prediction (every vessel labeled normal) scored directly against the ground truth. VS-$F_1$ (Vessel Severity Macro-$F_1$) is the primary diagnostic metric, reported in 2-class (lesion/non-lesion) and 3-class (normal/mild--moderate/severe) forms. ROUGE-L and BERTScore are supplementary reference-based text-similarity metrics whose scores need not solely reflect diagnostic agreement (Section~\ref{sec:emergence}). Stage~3 (RLVR w/o Think) and CARDEA (Final) use the same Stage~3 model with thinking disabled and enabled, respectively.\par
Across five independent MedGemma label-extraction runs over the same fixed reports (sampling temperature 0.5), the standard deviation of VS-$F_1$ was below 0.003 for every entry, indicating that label extraction is stable under stochastic decoding.\par}
\end{table*}

\subsection{Ablation Studies}
\label{sec:ablations}

CoB reasoning is not native to the base Qwen3-VL. Against the selected configuration (cold start with a conditional CoB reward), we compared three variants, each altering one design choice: no cold start, no reward, or an unconditional reward (Table~\ref{tab:ablations}).

\begin{figure*}[t]
\centering
\begin{minipage}{0.48\textwidth}
  \centering
  \includegraphics[width=\linewidth]{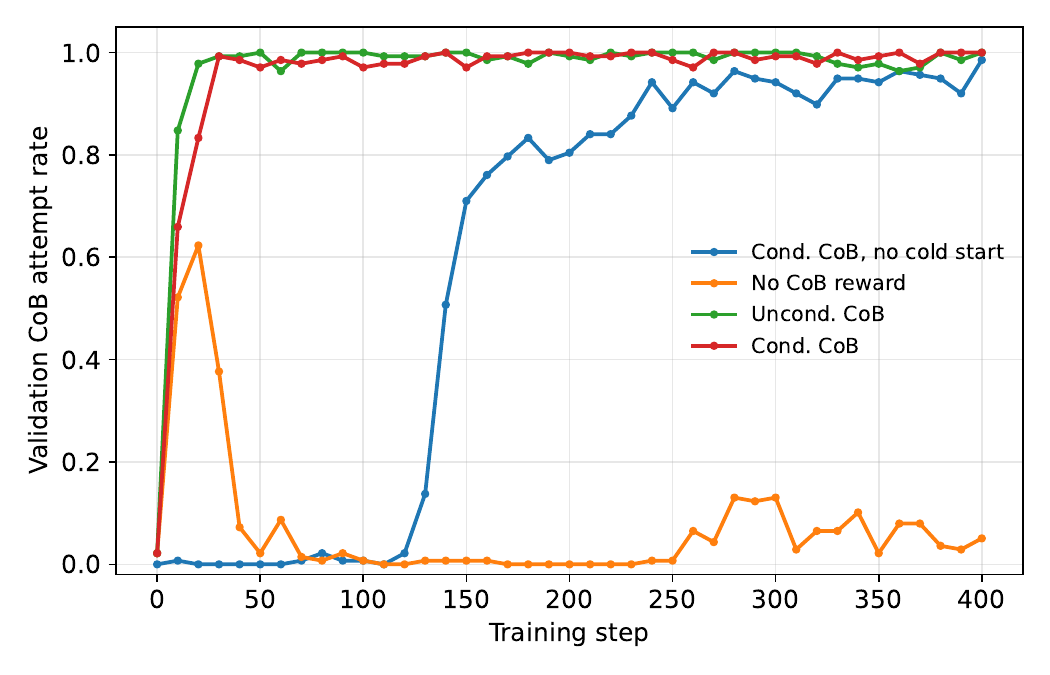}
  \vspace{0.5ex}
  \centerline{\textbf{A}}
\end{minipage}
\hfill
\begin{minipage}{0.48\textwidth}
  \centering
  \includegraphics[width=\linewidth]{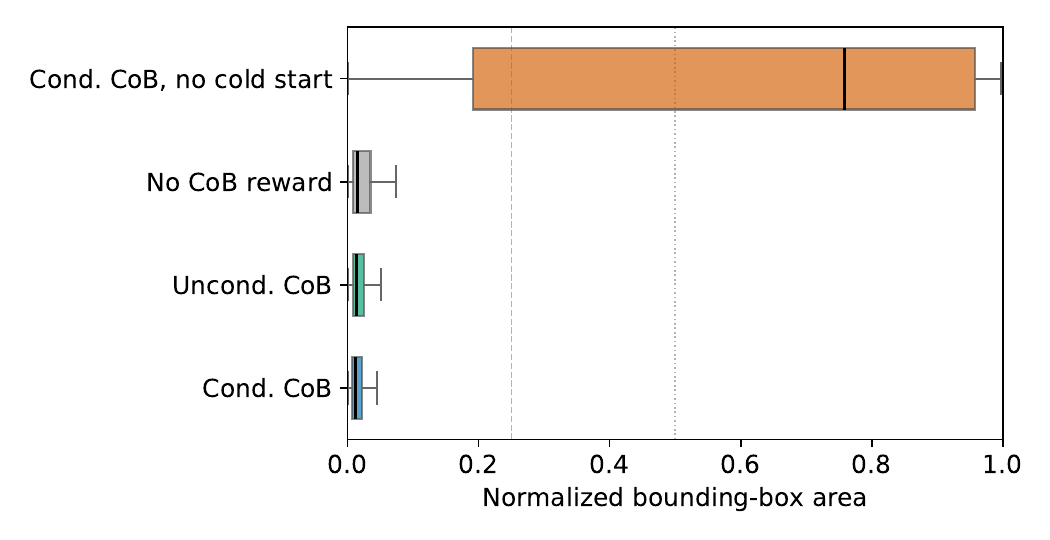}
  \vspace{0.5ex}
  \centerline{\textbf{B}}
\end{minipage}
\caption{Ablation of cold-start initialization and the CoB reward. \textbf{A,} Validation CoB usage across RLVR training. Without a CoB reward, usage briefly rises then collapses to $\sim$5\%; both cold-start configurations climb to near 1.0 within $\sim$30 steps; without the cold start, usage stays near zero for $\sim$130 steps before rising. \textbf{B,} Distribution of in-trace box areas on the held-out test classification tasks (box-and-whisker, normalized to full-frame area). Without the cold start, boxes degenerate to near-whole-frame; all cold-start variants emit compact, vessel-scale boxes.}
\label{fig:ablation_grounding}
\end{figure*}

Removing the cold start slowed CoB adoption (about 130 steps to near-full usage vs.\ about 30 with it; Figure~\ref{fig:ablation_grounding}A) but was not decisive. The reward eventually pulled usage to the same level, and accuracy stayed comparable. The cold start's real effect was on box scale, as its traces inherit the fine-grained vessel and stenosis boxes distilled from Stage~1. The reward asks only for a box and is silent on its size, so without that demonstration the model keeps whatever coarse boxes still earn the reward. In our runs these spanned nearly the whole frame (median 0.76 of the frame area vs.\ about 0.01 with a cold start; Figure~\ref{fig:ablation_grounding}B), running counter to the interpretability purpose.

Removing the CoB reward let grounding collapse to about 5\% (Figure~\ref{fig:ablation_grounding}A), yet accuracy stayed comparable. CoB is therefore not what drives accuracy; rewarding it does no harm and is what sustains the grounding.

Making the reward unconditional left accuracy statistically indistinguishable on our test splits. In a related tool-use setting, DeepEyes~\cite{zheng2026deepeyes} reported a clearer gap, with the conditional reward converging to higher accuracy. We nonetheless retain the conditional form: it had the highest validation point estimates and is the stricter rule, never rewarding a box on a wrong answer.

\begin{table}[htbp]
\centering
\caption{Ablation over the Cold Start and the CoB Reward. Bold marks the last column, the configuration selected for CARDEA.}
\label{tab:ablations}
\small
\setlength{\tabcolsep}{6pt}
\renewcommand{\arraystretch}{1.15}
\resizebox{\textwidth}{!}{%
\begin{tabular}{l cccc}
\toprule
\textbf{RLVR initialization} & Align & Cold start & Cold start & \textbf{Cold start} \\
\textbf{CoB reward} & Cond. & None & Uncond. & \textbf{Cond.} \\
\midrule
\multicolumn{5}{l}{\textit{Validation}} \\
\quad Dominance & 0.80 & 0.82 & 0.77 & 0.84 \\
\quad Complexity Assess. & 0.71 & 0.75 & 0.71 & 0.76 \\
\midrule
\multicolumn{5}{l}{\textit{Test}} \\
\quad Dominance (Real)  & 0.81 (0.75--0.86) & 0.81 (0.75--0.86) & 0.73 (0.68--0.79) & 0.76 (0.70--0.81) \\
\quad Dominance (Shift) & 0.78 (0.71--0.85) & 0.86 (0.80--0.91) & 0.79 (0.72--0.86) & 0.86 (0.80--0.91) \\
\quad Complexity Assess. & 0.73 (0.60--0.84) & 0.82 (0.69--0.92) & 0.81 (0.69--0.91) & 0.85 (0.73--0.94) \\
\quad Report (2-class) & 0.661 (0.633--0.689) & 0.613 (0.587--0.639) & 0.665 (0.640--0.691) & 0.680 (0.654--0.706) \\
\bottomrule
\end{tabular}%
}

\vspace{0.5ex}
{\footnotesize\setstretch{1}\raggedright
Test entries are point estimates with 95\% confidence intervals in parentheses, obtained by recomputing each metric across 5{,}000 case-level bootstrap resamples. Validation entries are monitor-only and are shown as point estimates alone.\par
The four variants were each trained for 400 steps under a fixed budget and are distinct from the full CARDEA model. Validation rows are used for model selection and are the mean over the final five validation checkpoints (steps 360--400, evaluated every 10 steps), from a single run and on different populations from the test rows. Test rows are the official CoronaryDominance/CardioSyntax test splits and zero-shot AngioCAD report generation.\par
Align denotes initialization from the Stage~1 aligned model and Cold start from the Stage~2 cold-start model; Cond., conditional CoB reward; Uncond., unconditional CoB reward; and None, no CoB reward. Real and Shift denote the Real Distribution and Domain Shift test sets, respectively. Dominance classification and complexity assessment use Macro $F_1$; report generation uses the two-class vessel-severity macro-$F_1$.\par}
\end{table}

\section{Discussion}
\label{sec:discussion}

CARDEA is a generalist that broadly matches its specialized baselines, but its aim is not to win every subtask: it is to be the single inference core of a two-pass pipeline that delivers study-level diagnoses and grounds them in auditable bounding boxes within the reasoning trace, which no prior CAG system does \cite{avram2023cathai, labrecque2024evaluation, harrabi2026deepcoroclip, nakamura2025cag, jiang2026vision}. The pipeline's final diagnosis lies in the study-level tasks, where it matched two cardiologists' accuracy on complexity assessment and drew level with the dedicated classifier under domain shift, its only distinguishable shortfall being Real Distribution dominance. Its view filtering also raised all three zero-shot point estimates, suggesting view completeness improves information density beyond its computational savings.

Jiang et al.~\cite{jiang2026vision} supervised report generation directly and it still performed poorly, which they attribute to pairing one comprehensive report with no intermediate reasoning to link findings to statements. Our results suggest a different route: report generation was excluded from training, yet the two supervised stages eroded its zero-shot quality, and only RLVR reversed the decline. We hypothesize that each verified answer forces CARDEA to run its own multi-view synthesis, which the reward repeatedly refines; the RLVR model surpasses the untuned base even with native thinking disabled, indicating this synthesis is internalized rather than emitted in the trace. By contrast, imitation copies only the output form, consistent with the decline under supervision. Although this interpretation remains hypothetical and rests on a single run, it echoes DeepSeek-R1~\cite{deepseek2025r1}, where policy optimization surfaced capabilities that imitation did not reach.

CARDEA's auditable grounding rests on three design choices. The cold-start distillation drives box quality: without it, boxes tend to be coarse and less interpretable (Figure~\ref{fig:ablation_grounding}B). The CoB reward sustains the grounding at no cost to accuracy: on every split, each rewarded configuration remained comparable to or above the variant trained without it (Table~\ref{tab:ablations}). Conditioning that reward on a correct answer yielded higher validation point estimates than the unconditional form, but the two forms remained comparable at test.

We also weigh the trade-offs CARDEA would face in deployment. First, its autoregressive architecture yields a median inference time of about 6 seconds per study even under tensor parallelism on $4\times$NVIDIA~B200 GPUs (Appendix~\ref{sec:appendix_extended_limitations}), so it cannot serve applications that need a sub-second response. Second, although CARDEA performed CoB on nearly every in-distribution diagnosis (usage 99.5\% on Real Distribution dominance and 100\% on complexity assessment), its usage falls on out-of-distribution imaging or tasks (65\% on Domain Shift dominance and 75\% on report generation). Repeated sampling narrowed both gaps, raising Domain Shift dominance usage from 65\% to 92\% and report-generation usage from 75\% to 98.5\% across seven rollouts without materially changing task performance (Appendix~\ref{sec:appendix_cob_coverage}). Deployment should therefore be tested against the expected data distribution first, and closing the gap fully may require further SFT on rejection-sampled trajectories from out-of-distribution data.

These conclusions are subject to several limitations. Above all, CARDEA has not been validated in a clinical setting, and its reports have not undergone blinded comparison with physician-authored reports; prospective validation against expert cardiologists is a prerequisite for clinical use. Training limitations include cross-stage comparisons based on one run per stage; training data focused mainly on vessels and stenoses, leaving other findings insufficiently learned; and study-level training using at most 10 static keyframes per study without the temporal dynamics of contrast flow. Evaluation limitations include CoB assessment based only on box frequency, box area, and final-answer correctness, not box faithfulness; the report metric's lack of a human-expert reference, requiring report scores to be interpreted relative to the naive floor and across stages rather than as absolute measures of report quality; and two small test sets (complexity assessment, $n{=}60$; keyframe selection, 48 videos from 5 patients) that may be underpowered. Additional limitations and technical considerations are discussed in Appendix~\ref{sec:appendix_extended_limitations}.

\section{Conclusion}
\label{sec:conclusion}

We presented CARDEA, a unified LVLM for end-to-end CAG interpretation developed without human-annotated reasoning traces. Its central contribution is an auditable decision process in which CoB evidence links conclusions to specific regions in the source images, allowing clinicians to inspect the spatial basis of each diagnosis. After three-stage training on closed-ended tasks, CARDEA remained competitive with task-specific baselines and performed comparably to two interventional cardiologists on complexity assessment. It also generalized zero-shot to the fully held-out AngioCAD cohort, surpassing the published baseline for RCA stenosis classification. Although report generation was excluded from training, the RLVR stage reversed the decline under SFT and raised zero-shot performance above the untuned base model. Together, these results establish the feasibility of auditable end-to-end CAG interpretation but do not yet establish readiness for clinical use. Prospective validation against expert cardiologists remains essential before deployment.

\section*{Ethics Approval}

The research protocol encompassing this work was approved by the Research Ethics Committee of China Medical University and Hospital, Taichung, Taiwan (IRB No. CMUH114-REC2-012), with a waiver of informed consent. All analyses reported in this manuscript used only publicly available, de-identified datasets; no patient data from China Medical University Hospital were analyzed.

\section*{Funding}

This work was supported by the National Science and Technology Council, Taiwan (Grant No. NSTC 114-2314-B-039-073), and China Medical University Hospital (Grant Nos. DMR-115-087 and DMR-115-108).

\section*{Use of Artificial Intelligence}

This study used ChatGPT (OpenAI) and Claude (Anthropic) to improve the manuscript's language, grammar, and style. The authors conceived and drafted the manuscript. These tools also assisted in debugging and optimizing portions of the experimental code. The authors reviewed all AI-generated suggestions and finalized all modifications. They assume full responsibility for the scientific accuracy and integrity of the work. No protected health information was used or disclosed because all datasets were publicly available and de-identified.

\section*{Data Sharing Statement}
All datasets used in this study are publicly available \cite{jimenez2024cadica, popov2024dataset, chen2024huatuogpt, kruzhilov2025coronarydominance, ponomarchuk2025cardiosyntax, hosseini2026angiocad}. The model weights are available at \url{https://huggingface.co/benbayibaurba/cardea-v0}, and the inference code at \url{https://github.com/benbayibaurba/cardea}.

\bibliographystyle{unsrtnat}
\bibliography{references}

@article{deepseek2025r1,
  title={Deepseek-r1: Incentivizing reasoning capability in llms via reinforcement learning},
  author={Guo, Daya and Yang, Dejian and Zhang, Haowei and Song, Junxiao and Zhang, Ruoyu and Xu, Runxin and Zhu, Qihao and Ma, Shirong and Wang, Peiyi and Bi, Xiao and others},
  journal={arXiv preprint arXiv:2501.12948},
  year={2025}
}

@article{gao2025soft,
  title={Soft adaptive policy optimization},
  author={Gao, Chang and Zheng, Chujie and Chen, Xiong-Hui and Dang, Kai and Liu, Shixuan and Yu, Bowen and Yang, An and Bai, Shuai and Zhou, Jingren and Lin, Junyang},
  journal={arXiv preprint arXiv:2511.20347},
  year={2025}
}

@misc{bai2025qwen3vltechnicalreport,
      title={Qwen3-VL Technical Report}, 
      author={Shuai Bai and Yuxuan Cai and Ruizhe Chen and Keqin Chen and Xionghui Chen and Zesen Cheng and Lianghao Deng and Wei Ding and Chang Gao and Chunjiang Ge and Wenbin Ge and Zhifang Guo and Qidong Huang and Jie Huang and Fei Huang and Binyuan Hui and Shutong Jiang and Zhaohai Li and Mingsheng Li and Mei Li and Kaixin Li and Zicheng Lin and Junyang Lin and Xuejing Liu and Jiawei Liu and Chenglong Liu and Yang Liu and Dayiheng Liu and Shixuan Liu and Dunjie Lu and Ruilin Luo and Chenxu Lv and Rui Men and Lingchen Meng and Xuancheng Ren and Xingzhang Ren and Sibo Song and Yuchong Sun and Jun Tang and Jianhong Tu and Jianqiang Wan and Peng Wang and Pengfei Wang and Qiuyue Wang and Yuxuan Wang and Tianbao Xie and Yiheng Xu and Haiyang Xu and Jin Xu and Zhibo Yang and Mingkun Yang and Jianxin Yang and An Yang and Bowen Yu and Fei Zhang and Hang Zhang and Xi Zhang and Bo Zheng and Humen Zhong and Jingren Zhou and Fan Zhou and Jing Zhou and Yuanzhi Zhu and Ke Zhu},
      year={2025},
      eprint={2511.21631},
      archivePrefix={arXiv},
      primaryClass={cs.CV},
      url={https://arxiv.org/abs/2511.21631}, 
}

@article{jimenez2024cadica,
  title={CADICA: A new dataset for coronary artery disease detection by using invasive coronary angiography},
  author={Jim{\'e}nez-Partinen, Ariadna and Molina-Cabello, Miguel A and Thurnhofer-Hemsi, Karl and Palomo, Esteban J and Rodr{\'\i}guez-Capit{\'a}n, Jorge and Molina-Ramos, Ana I and Jim{\'e}nez-Navarro, Manuel},
  journal={Expert Systems},
  volume={41},
  number={12},
  pages={e13708},
  year={2024},
  publisher={Wiley Online Library}
}

@article{popov2024dataset,
  title={Dataset for automatic region-based coronary artery disease diagnostics using X-ray angiography images},
  author={Popov, Maxim and Amanturdieva, Akmaral and Zhaksylyk, Nuren and Alkanov, Alsabir and Saniyazbekov, Adilbek and Aimyshev, Temirgali and Ismailov, Eldar and Bulegenov, Ablay and Kuzhukeyev, Arystan and Kulanbayeva, Aizhan and others},
  journal={Scientific data},
  volume={11},
  number={1},
  pages={20},
  year={2024},
  publisher={Nature Publishing Group UK London}
}

@inproceedings{ponomarchuk2025cardiosyntax,
  title={CardioSyntax: End-to-End SYNTAX Score Prediction-Dataset, Benchmark and Method},
  author={Ponomarchuk, Alexander and Kruzhilov, Ivan and Mazanov, Gleb and Utegenov, Ruslan and Shadrin, Artem and Zubkova, Galina and Bessonov, Ivan and Blinov, Pavel},
  booktitle={2025 IEEE/CVF Winter Conference on Applications of Computer Vision (WACV)},
  pages={5873--5883},
  year={2025},
  organization={IEEE}
}

@article{kruzhilov2025coronarydominance,
  title={CoronaryDominance: Angiogram dataset for coronary dominance classification},
  author={Kruzhilov, Ivan and Mazanov, Gleb and Ponomarchuk, Alexander and Zubkova, Galina and Shadrin, Artem and Utegenov, Ruslan and Blinov, Pavel and Bessonov, Ivan},
  journal={Scientific Data},
  volume={12},
  number={1},
  pages={341},
  year={2025},
  publisher={Nature Publishing Group UK London}
}

@article{chen2024huatuogpt,
  title={Huatuogpt-vision, towards injecting medical visual knowledge into multimodal llms at scale},
  author={Chen, Junying and Gui, Chi and Ouyang, Ruyi and Gao, Anningzhe and Chen, Shunian and Chen, Guiming Hardy and Wang, Xidong and Zhang, Ruifei and Cai, Zhenyang and Ji, Ke and others},
  journal={arXiv preprint arXiv:2406.19280},
  year={2024}
}

@article{bannur2024maira,
  title={Maira-2: Grounded radiology report generation},
  author={Bannur, Shruthi and Bouzid, Kenza and Castro, Daniel C and Schwaighofer, Anton and Thieme, Anja and Bond-Taylor, Sam and Ilse, Maximilian and P{\'e}rez-Garc{\'\i}a, Fernando and Salvatelli, Valentina and Sharma, Harshita and others},
  journal={arXiv preprint arXiv:2406.04449},
  year={2024}
}

@article{nakamura2025cag,
  title={CAG-VLM: Fine-Tuning of a Large-Scale Model to Recognize Angiographic Images for Next-Generation Diagnostic Systems},
  author={Nakamura, Yuto and Kodera, Satoshi and Settai, Haruki and Shinohara, Hiroki and Tamura, Masatsugu and Noguchi, Tomohiro and Furusawa, Tatsuki and Takizawa, Ryo and Kabayama, Tempei and Takeda, Norihiko},
  journal={arXiv preprint arXiv:2505.04964},
  year={2025}
}

@article{avram2023cathai,
  title={CathAI: fully automated coronary angiography interpretation and stenosis estimation},
  author={Avram, Robert and Olgin, Jeffrey E and Ahmed, Zeeshan and Verreault-Julien, Louis and Wan, Alvin and Barrios, Joshua and Abreau, Sean and Wan, Derek and Gonzalez, Joseph E and Tardif, Jean-Claude and others},
  journal={npj Digital Medicine},
  volume={6},
  number={1},
  pages={142},
  year={2023},
  publisher={Nature Publishing Group UK London}
}

@article{labrecque2024evaluation,
  title={Evaluation of stenoses using AI video models applied to coronary angiography},
  author={Labrecque Langlais, {\'E}lodie and Corbin, Denis and Tastet, Olivier and Hayek, Ahmad and Doolub, Gemina and Mrad, Sebasti{\'a}n and Tardif, Jean-Claude and Tanguay, Jean-Fran{\c{c}}ois and Marquis-Gravel, Guillaume and Tison, Geoffrey H and others},
  journal={NPJ digital medicine},
  volume={7},
  number={1},
  pages={138},
  year={2024},
  publisher={Nature Publishing Group UK London}
}

@article{sianos2005syntax,
  title={The SYNTAX Score: an angiographic tool grading the complexity of coronary artery disease},
  author={Sianos, Georgios and Morel, Marie-Ang{\`e}le and Kappetein, Arie Pieter and Morice, Marie-Claude and Colombo, Antonio and Dawkins, Keith and Van Den Brand, Marcel and Van Dyck, Nic and Russell, Mary E and Mohr, Friedrich W and others},
  journal={EuroIntervention},
  volume={1},
  number={2},
  pages={219--227},
  year={2005}
}

@article{li2023llava,
  title={Llava-med: Training a large language-and-vision assistant for biomedicine in one day},
  author={Li, Chunyuan and Wong, Cliff and Zhang, Sheng and Usuyama, Naoto and Liu, Haotian and Yang, Jianwei and Naumann, Tristan and Poon, Hoifung and Gao, Jianfeng},
  journal={Advances in Neural Information Processing Systems},
  volume={36},
  pages={28541--28564},
  year={2023}
}

@inproceedings{xu2025medground,
  title={Medground-r1: Advancing medical image grounding via spatial-semantic rewarded group relative policy optimization},
  author={Xu, Huihui and Nie, Yuanpeng and Wang, Hualiang and Chen, Ying and Li, Wei and Ning, Junzhi and Liu, Lihao and Wang, Hongqiu and Zhu, Lei and Liu, Jiyao and others},
  booktitle={International Conference on Medical Image Computing and Computer-Assisted Intervention},
  pages={391--401},
  year={2025},
  organization={Springer}
}

@article{fan2025grit,
  title={GRIT: Teaching MLLMs to Think with Images},
  author={Fan, Yue and He, Xuehai and Yang, Diji and Zheng, Kaizhi and Kuo, Ching-Chen and Zheng, Yuting and Narayanaraju, Sravana Jyothi and Guan, Xinze and Wang, Xin Eric},
  journal={arXiv preprint arXiv:2505.15879},
  year={2025}
}

@inproceedings{zheng2026deepeyes,
  title={DeepEyes: Incentivizing ``Thinking with Images'' via Reinforcement Learning},
  author={Zheng, Ziwei and Yang, Minghao and Hong, Jack and Zhao, Chenxiao and Xu, Guohai and Yang, Le and Shen, Chao and Yu, Xing},
  booktitle={International Conference on Learning Representations},
  volume={2026},
  pages={126775--126798},
  year={2026},
  url={https://proceedings.iclr.cc/paper_files/paper/2026/file/cdb347d7516d52a9280ea9d6708911f6-Paper-Conference.pdf}
}

@inproceedings{yildirim2024multimodal,
  title={Multimodal healthcare AI: identifying and designing clinically relevant vision-language applications for radiology},
  author={Yildirim, Nur and Richardson, Hannah and Wetscherek, Maria Teodora and Bajwa, Junaid and Jacob, Joseph and Pinnock, Mark Ames and Harris, Stephen and Coelho De Castro, Daniel and Bannur, Shruthi and Hyland, Stephanie and others},
  booktitle={Proceedings of the 2024 CHI Conference on Human Factors in Computing Systems},
  pages={1--22},
  year={2024}
}

@article{moon2021automatic,
  title={Automatic stenosis recognition from coronary angiography using convolutional neural networks},
  author={Moon, Jong Hak and Cha, Won Chul and Chung, Myung Jin and Lee, Kyu-Sung and Cho, Baek Hwan and Choi, Jin Ho and others},
  journal={Computer methods and programs in biomedicine},
  volume={198},
  pages={105819},
  year={2021},
  publisher={Elsevier}
}

@article{sellergren2025medgemma,
  title={Medgemma technical report},
  author={Sellergren, Andrew and Kazemzadeh, Sahar and Jaroensri, Tiam and Kiraly, Atilla and Traverse, Madeleine and Kohlberger, Timo and Xu, Shawn and Jamil, Fayaz and Hughes, C{\'\i}an and Lau, Charles and others},
  journal={arXiv preprint arXiv:2507.05201},
  year={2025}
}

@article{harrabi2026deepcoroclip,
  title={DeepCORO-CLIP: A Multi-View Foundation Model for Comprehensive Coronary Angiography Video-Text Analysis and External Validation},
  author={Harrabi, Sarra and Wu, Y and Vukadinovic, M and others},
  journal={arXiv preprint arXiv:2603.17675},
  year={2026}
}

@article{chen2025computational,
  title={Computational Methods for Analysing X-ray-guided Coronary Angiography},
  author={Chen, Tiana and Yap, Jonathan and Keong, Yeo Khung and others},
  journal={Journal of Asian Pacific Society of Cardiology (JAPSC)},
  volume={4},
  pages={e12},
  year={2025}
}

@article{pascual2025hyperparameter,
  title={Hyperparameter optimization of YOLO models for invasive coronary angiography lesion detection and assessment},
  author={Pascual-Gonz{\'a}lez, Mario and Jim{\'e}nez-Partinen, Ariadna and Palomo, Esteban J and L{\'o}pez-Rubio, Ezequiel and Ortega-G{\'o}mez, Almudena},
  journal={Computers in Biology and Medicine},
  volume={196},
  pages={110697},
  year={2025},
  publisher={Elsevier}
}

@article{jiang2026vision,
  title={Vision Language Model for Coronary Angiogram Analysis and Report Generation: Development and Evaluation Study},
  author={Jiang, Qianfeng and Ke, Yuhe and Sinisterra, Laura Gutierrez and Elangovan, Kabilan and Li, Zengxiang and Yeo, Khung Keong and Jonathan, Yap and Ting, Daniel Shu Wei},
  journal={medRxiv},
  pages={2026--04},
  year={2026},
  publisher={Cold Spring Harbor Laboratory Press}
}

@article{chen2025metis,
  title={Metis-SPECS: Decoupling Multimodal Learning via Self-distilled Preference-based Cold Start},
  author={Chen, Kun and Shi, Peng and Qiu, Haibo and Zeng, Zhixiong and Yang, Siqi and Mao, Wenji and Ma, Lin},
  journal={arXiv preprint arXiv:2510.25801},
  year={2025}
}

@article{hosseini2026angiocad,
  title={AngioCAD: A Public X-Ray Angiography Dataset and an Adaptive Fusion Framework for Stenosis Detection},
  author={Hosseini, Marzieh Sadat and Naghsh-Nilchi, Ahmad R and Safayani, Mehran and Sadeghi, Masoumeh and Shirvani, Ehsan and Danesh, Manizheh and Miramirkhani, Seyed Ali},
  journal={Computer Methods and Programs in Biomedicine},
  pages={109368},
  year={2026},
  publisher={Elsevier}
}

@article{moalla2023deep,
  title={Deep AngioKey: A Novel Approach for Objective Keyframe Extraction in Coronary Angiography Analysis},
  author={Moalla, Hounaida and Ghrab, Aiman and Bahloul, Amine and Hamed, Bassem Ben and Abid, Leila},
  journal={Research Square preprint},
  doi={10.21203/rs.3.rs-3498564/v1},
  year={2023}
}

@inproceedings{lin2004rouge,
  title={Rouge: A package for automatic evaluation of summaries},
  author={Lin, Chin-Yew},
  booktitle={Text summarization branches out},
  pages={74--81},
  year={2004}
}

@article{zhang2019bertscore,
  title={Bertscore: Evaluating text generation with bert},
  author={Zhang, Tianyi and Kishore, Varsha and Wu, Felix and Weinberger, Kilian Q and Artzi, Yoav},
  journal={arXiv preprint arXiv:1904.09675},
  year={2019}
}

@article{tran2025anatomy,
  title={Anatomy-specific two-stage YOLOv8 approach for improved coronary segmentation using the ARCADE dataset},
  author={Tran, Dinh-Son and Huynh, Anh-Khoa and Huynh, Anh-Duy and Nguyen-Thoi, Trung},
  journal={Optics Continuum},
  volume={4},
  number={2},
  pages={303--317},
  year={2025},
  publisher={Optica Publishing Group}
}

@inproceedings{zheng2024llamafactory,
  title={LlamaFactory: Unified Efficient Fine-Tuning of 100+ Language Models},
  author={Yaowei Zheng and Richong Zhang and Junhao Zhang and Yanhan Ye and Zheyan Luo and Zhangchi Feng and Yongqiang Ma},
  booktitle={Proceedings of the 62nd Annual Meeting of the Association for Computational Linguistics (Volume 3: System Demonstrations)},
  address={Bangkok, Thailand},
  publisher={Association for Computational Linguistics},
  year={2024},
  url={http://arxiv.org/abs/2403.13372}
}

@misc{zheng2025easyr1,
  title        = {EasyR1: An Efficient, Scalable, Multi-Modality RL Training Framework},
  author       = {Yaowei Zheng and Junting Lu and Shenzhi Wang and Zhangchi Feng and Dongdong Kuang and Yuwen Xiong and Richong Zhang},
  howpublished = {\url{https://github.com/hiyouga/EasyR1}},
  year         = {2025}
}

@article{sheng2024hybridflow,
  title   = {HybridFlow: A Flexible and Efficient RLHF Framework},
  author  = {Guangming Sheng and Chi Zhang and Zilingfeng Ye and Xibin Wu and Wang Zhang and Ru Zhang and Yanghua Peng and Haibin Lin and Chuan Wu},
  year    = {2024},
  journal = {arXiv preprint arXiv:2409.19256}
}

@article{hu2021lora,
  title={Lora: Low-rank adaptation of large language models},
  author={Hu, Edward J and Shen, Yelong and Wallis, Phillip and Allen-Zhu, Zeyuan and Li, Yuanzhi and Wang, Shean and Wang, Lu and Chen, Weizhu},
  journal={arXiv preprint arXiv:2106.09685},
  year={2021}
}

@article{loshchilov2017decoupled,
  title={Decoupled weight decay regularization},
  author={Loshchilov, Ilya and Hutter, Frank},
  journal={arXiv preprint arXiv:1711.05101},
  year={2017}
}

@inproceedings{rajbhandari2020zero,
  title={Zero: Memory optimizations toward training trillion parameter models},
  author={Rajbhandari, Samyam and Rasley, Jeff and Ruwase, Olatunji and He, Yuxiong},
  booktitle={SC20: international conference for high performance computing, networking, storage and analysis},
  pages={1--16},
  year={2020},
  organization={IEEE}
}

@article{zhao2023pytorch,
  title={Pytorch fsdp: experiences on scaling fully sharded data parallel},
  author={Zhao, Yanli and Gu, Andrew and Varma, Rohan and Luo, Liang and Huang, Chien-Chin and Xu, Min and Wright, Less and Shojanazeri, Hamid and Ott, Myle and Shleifer, Sam and others},
  journal={arXiv preprint arXiv:2304.11277},
  year={2023}
}

@inproceedings{kwon2023efficient,
  title={Efficient Memory Management for Large Language Model Serving with PagedAttention},
  author={Woosuk Kwon and Zhuohan Li and Siyuan Zhuang and Ying Sheng and Lianmin Zheng and Cody Hao Yu and Joseph E. Gonzalez and Hao Zhang and Ion Stoica},
  booktitle={Proceedings of the ACM SIGOPS 29th Symposium on Operating Systems Principles},
  year={2023}
}

@article{oikonomou2022current,
  title={Current Concepts and Future Applications of Non-Invasive Functional and Anatomical Evaluation of Coronary Artery Disease},
  author={Oikonomou, Evangelos and Theofilis, Panagiotis and Lampsas, Stamatios and Katsarou, Ourania and Kalogeras, Konstantinos and Marinos, Georgios and Tsatsaragkou, Aikaterini and Anastasiou, Artemis and Lysandrou, Antonios and Gounaridi, Maria-Ioanna and Gialamas, Ioannis and Vavuranakis, Michael-Andrew and Tousoulis, Dimitris and Vavuranakis, Manolis and Siasos, Gerasimos},
  journal={Life},
  volume={12},
  number={11},
  pages={1803},
  year={2022},
  doi={10.3390/life12111803}
}

@article{cimen2016reconstruction,
  title={Reconstruction of Coronary Arteries from X-Ray Angiography: A Review},
  author={{\c{C}}imen, Serkan and Gooya, Ali and Grass, Michael and Frangi, Alejandro F.},
  journal={Medical Image Analysis},
  volume={32},
  pages={46--68},
  year={2016},
  doi={10.1016/j.media.2016.02.007}
}

@article{green2016optimal,
  title={Optimal Angiographic Views for Invasive Coronary Angiography: A Guide for Trainees},
  author={Green, Peregrine and Frobisher, Paul and Ramcharitar, Steve},
  journal={British Journal of Cardiology},
  volume={23},
  pages={110--113},
  year={2016},
  doi={10.5837/bjc.2016.028}
}

@article{shivaie2024interobserver,
  title={Interobserver Variability of Coronary Stenosis Characterized by Coronary Angiography: A Single-Center (Toronto General Hospital) Retrospective Chart Review by Staff Cardiologists},
  author={Shivaie, Seyedmohammadshahab and Tohidi, Hadi and Loganathan, Pragash and Kar, Manish and Hashemy, Habiba and Shafiee, Mohammad A.},
  journal={Vascular Health and Risk Management},
  volume={20},
  pages={359--368},
  year={2024},
  doi={10.2147/VHRM.S431612}
}

\clearpage
\appendix
\renewcommand{\thesection}{S\arabic{section}}
\setcounter{section}{0}
\renewcommand{\thetable}{S\arabic{table}}
\setcounter{table}{0}
\renewcommand{\thefigure}{S\arabic{figure}}
\setcounter{figure}{0}
\renewcommand{\theequation}{S\arabic{equation}}
\setcounter{equation}{0}
\renewcommand{\theHtable}{S\arabic{table}}
\renewcommand{\theHfigure}{S\arabic{figure}}
\renewcommand{\theHequation}{S\arabic{equation}}
\renewcommand{\theHsection}{S\arabic{section}}
\renewcommand{\theHsubsection}{S\arabic{section}.\arabic{subsection}}
\begin{center}
{\large\textbf{Supplementary Appendix}}
\end{center}
\section{Supplementary Methods}
\label{sec:appendix_metrics}

\subsection{Datasets and Preprocessing}
\label{sec:appendix_datasets}
CARDEA is designed to perform both single-view and study-level reasoning within our end-to-end pipeline. We categorize the datasets accordingly and describe the preprocessing and derived tasks for each.

\subsubsection{Single-View Datasets}
\begin{itemize}
    \item \textbf{ARCADE \cite{popov2024dataset}:} A dataset providing 3,000 single-view keyframes compiled from 1,500 distinct patients using diverse imaging equipment. It includes expert annotations for 25 distinct coronary segments (defined by the SYNTAX score) and localizations for stenotic lesions ($\ge 50\%$ diameter stenosis). We utilize these annotations to construct three tasks: (1) \textit{Vessel Detection}, outputting spatial bounding boxes for specific vessel segments; (2) \textit{Stenosis Detection}, outputting bounding boxes for stenotic lesions; and (3) \textit{View Classification} for left coronary artery (LCA), right coronary artery (RCA), or OTHER. The LCA and RCA labels are derived from the segment annotations. To formulate the ``OTHER'' class, we manually curated a small set of non-coronary-angiographic medical images from PubMedVision \cite{chen2024huatuogpt} alongside non-diagnostic, low-quality frames from CADICA \cite{jimenez2024cadica}, enabling the model to train for view recognition and image usability simultaneously; this patient diversity supports generalization. The segment and stenosis annotations form two separate 1,500-image subsets, each with its own official split of 1,000 training, 200 validation, and 300 test samples.

    \item \textbf{CADICA \cite{jimenez2024cadica}:} A dataset comprising 668 multi-view angiographic videos from 42 patients, among which 382 videos were expertly curated as diagnostic. Inspired by Jiang et al. \cite{jiang2026vision}, who used CADICA to train a standalone ViT-based frame selector, we use these expert-labeled frames to train our unified LVLM for the \textit{Keyframe Selection} task. As a preprocessing step, each frame is resized from $512\times512$ to $256\times256$; videos exceeding 50 frames are subsampled to 50 evenly spaced frames to bound computational cost. Beyond keyframe selection, CADICA also carries multi-level stenosis annotations (bounding boxes labeled across seven severity levels, from sub-50\% to total occlusion); we include these as an auxiliary localization signal during feature alignment, exposing the model to lesions milder than ARCADE's 50\% stenosis threshold. We do not benchmark this task in isolation. With only 42 patients, highly correlated intra-video frames, and labels split across seven severity levels, the per-level data is too limited to support a separate localization benchmark. Since no official split is provided, we partition the 42 patients by stratified sampling on maximum stenosis severity into 32 training, 5 validation, and 5 test patients.
\end{itemize}

\subsubsection{Preliminary LVLM for Study-Level Dataset Preparation}
\label{sec:preliminary_lvlm}
To construct the study-level datasets, we use the single-view tasks from ARCADE and CADICA to train a preliminary LVLM that serves purely as an offline data-preparation utility. For each raw video, it selects the optimal diagnostic keyframe, classifies it as LCA, RCA, or OTHER, discards non-diagnostic images, and ensures each study retains both valid LCA and RCA views. To bound the visual token budget, we cap each study at 10 keyframes. For the training and validation splits we additionally keep only those with at least four to maintain quality, whereas the official test sets are retained in full to preserve comparability with the published baselines. This preliminary LVLM is used only to build the training datasets (CoronaryDominance~\cite{kruzhilov2025coronarydominance} and CardioSyntax~\cite{ponomarchuk2025cardiosyntax}), as the final CARDEA model does not yet exist at that stage; on the held-out AngioCAD~\cite{hosseini2026angiocad}, all steps---keyframe selection, view filtering, and diagnostic reasoning---are instead performed end-to-end by the fully trained CARDEA, as in its intended deployment.

\subsubsection{Study-Level Datasets}
\begin{itemize}
    \item \textbf{CoronaryDominance \cite{kruzhilov2025coronarydominance}:} A dataset comprising 1,574 multi-view angiographic studies. Each study is strictly labeled as either Left or Right dominance according to the SYNTAX score definitions (co-dominance is excluded). This dataset is used for the study-level task of \textit{Dominance Classification}. After the offline preprocessing (Section~\ref{sec:preliminary_lvlm}), the Main subset yields 924 training and 100 validation studies. For evaluation, we utilize the two official independent test sets: the Real Distribution subset (400 studies reflecting clinical class imbalance) and the Domain Shift subset (149 studies acquired with distinct imaging equipment, an out-of-distribution cohort).

    \item \textbf{CardioSyntax \cite{ponomarchuk2025cardiosyntax}:} A dataset encompassing 1,844 angiographic studies with expert-evaluated continuous SYNTAX scores, officially split into 1,784 training and 60 test studies. Given the difficulty of predicting continuous SYNTAX scores from images alone, we formulate a binary \textit{Complexity Assessment} task by discretizing the scores into normal-to-intermediate complexity ($0$--$32$) and high complexity ($> 32$). After applying the same offline preprocessing as in Section~\ref{sec:preliminary_lvlm}, 1,433 studies remain for training and 38 for validation. Evaluation is conducted on the official independent test set of 60 studies, for which the dataset additionally provides labels from two other cardiologists that serve as expert baselines.

    \item \textbf{AngioCAD~\cite{hosseini2026angiocad}:} A dataset of 413 angiographic studies annotated with stenosis grades for each coronary segment (defined by the American Heart Association coronary segment model). Excluded from all training stages, it serves exclusively as a zero-shot evaluation benchmark. We derive three tasks: (1) \textit{Multi-Frame View Classification} (LCA vs.\ RCA): following the original AngioCAD setup, CARDEA's single-view inference extracts the keyframe and samples 5 consecutive frames to predict the view; (2) \textit{Multi-Frame RCA Binary Stenosis Classification} (Lesion vs.\ Non-lesion): using the same 5-frame sampling restricted to RCA views, with each clip labeled Lesion ($>0\%$ stenosis) or Non-lesion; and (3) \textit{Report Generation}: the model generates a JSON clinical report covering the four major branches (LM, LAD, LCX, RCA), each field a free-text description with explicit segment references and stenosis-percentage estimates. Of the 413 studies, one is missing from the released data,\footnote{The missing study is case~157.} leaving 412 available for evaluation. To examine the effect of the end-to-end view filtering, we evaluate all three tasks on two cohorts: all 412 studies, and a valid-views subset of 326 studies retaining both an LCA and an RCA view.
\end{itemize}

\subsection{Evaluation Metrics}
\label{sec:appendix_eval_metrics}

Standard classification metrics including Macro $F_1$, Micro $F_1$,
Accuracy, and Balanced Accuracy follow conventional definitions and
are not elaborated here.

\subsubsection{Keyframe Selection}
\label{sec:appendix_dk}

We evaluate keyframe selection using $D_k$, defined as the minimum
absolute index distance between the model's predicted optimal frame
$idx_{best}^{pred}$ and the set of expert-annotated diagnostically
usable frames $S_{gt}$:
\begin{equation}
D_k = \min_{j \in S_{gt}} \left| idx_{best}^{pred} - j \right|
\end{equation}
A lower $D_k$ indicates better performance, with a value of zero
indicating that the predicted frame falls within the expert-annotated
usable range. This metric adapts the frame distance metric proposed in
Deep AngioKey~\cite{moalla2023deep}, which computes the distance from a
single designated ground-truth frame. Our adaptation instead measures
the distance from the full set of annotated usable frames because
CADICA provides a usable frame set rather than a single optimal index.

\subsubsection{Detection}
\label{sec:appendix_f1iou}

We evaluate detection at the instance level using one-to-one
matching between predicted bounding boxes $B_{pred}$ and
ground-truth boxes $B_{gt}$. A predicted box $b$ and a ground-truth
box $g$ are eligible for matching if they have the same semantic
label and an Intersection over Union (IoU) of at least 0.5. The set
of eligible pairs is
\begin{equation}
E_{0.5}^{\text{IoU}} =
\left\{(b,g) \in B_{pred}\times B_{gt}\;\middle|\;
b_{label}=g_{label}, \quad
\operatorname{IoU}(b_{box},g_{box})\ge0.5\right\}.
\end{equation}
The eligible pairs in $E_{0.5}^{\text{IoU}}$ are matched one-to-one,
with each predicted and ground-truth box appearing in at most one
match. The resulting pairs form $M_{\text{IoU}}$.

The resulting true-positive (TP), false-positive (FP), and
false-negative (FN) counts are
\begin{equation}
TP = |M_{\text{IoU}}|, \qquad
FP = |B_{pred}| - TP, \qquad
FN = |B_{gt}| - TP.
\end{equation}
Precision, recall, and $F_1$ are then computed as
\begin{equation}
P = \frac{TP}{TP + FP}, \qquad
R = \frac{TP}{TP + FN}, \qquad
F_1 = 2\,\frac{P R}{P + R}.
\end{equation}
We denote this instance-level score as $F_1$@IoU$\ge$0.5.

However, an IoU threshold of 0.5 can understate stenosis localization performance on CAG~\cite{jiang2026vision} because the criterion depends strongly on box extent. Since stenosis extent can be difficult to annotate consistently owing to the interpretive difficulty and inter-observer variability noted in the Introduction, the same diffuse lesion may be represented as one large box or several smaller boxes. A prediction that correctly localizes a lesion but differs in extent can therefore fall below the threshold and be counted as both a false positive and a false negative.

We therefore use Jiang et al.'s predicted-area overlap criterion for
stenosis detection to facilitate comparison. For a predicted box $b$
and a ground-truth box $g$, intersection over prediction (IoP) is
defined as
\begin{equation}
\operatorname{IoP}(b,g) =
\frac{\operatorname{area}(b_{box} \cap g_{box})}{\operatorname{area}(b_{box})}.
\end{equation}
After the IoU-based matching is completed, $U_{pred}$ and $U_{gt}$
denote the sets of predicted and ground-truth boxes that remain
unmatched, respectively. The pairs eligible under the IoP criterion
are
\begin{equation}
E_{0.6}^{\text{IoP}} =
\left\{(b,g) \in U_{pred}\times U_{gt}\;\middle|\;
b_{label}=g_{label}, \quad
\operatorname{IoP}(b,g)\ge0.6\right\}.
\end{equation}
The eligible pairs in $E_{0.6}^{\text{IoP}}$ are matched one-to-one.
The resulting pairs form $M_{\text{IoP}}$.

The final counts under the relaxed criterion are
\begin{equation}
TP_{rel} = |M_{\text{IoU}}| + |M_{\text{IoP}}|, \qquad
FP_{rel} = |B_{pred}| - TP_{rel}, \qquad
FN_{rel} = |B_{gt}| - TP_{rel}.
\end{equation}
This relaxation is applied only to stenosis detection; substituting the relaxed counts into the definitions above yields $F_1$@(IoU$\ge$0.5 or IoP$\ge$0.6).

\subsubsection{Report Generation}
\label{sec:appendix_eval_vsf1}

We evaluate report generation using vessel-severity macro-$F_1$
(VS-$F_1$), a task-specific application of macro-$F_1$ to severity
labels for the major coronary branches. Figure~\ref{fig:report_metric_example}
provides a worked example of the evaluation representations described in
this section.

AngioCAD provides segment-level stenosis grades for 15 coronary
segments. For evaluation, these segments are grouped into four major
branches: LM; LAD (proximal, mid, and distal LAD and the first and
second diagonal branches); LCX (proximal, mid, and distal LCX and the
obtuse marginal); and RCA (proximal, mid, and distal RCA, the posterior
descending artery, and the posterolateral branch).

We consolidate AngioCAD's seven original segment-level categories at
two granularities. For the primary two-class evaluation, the categories
are consolidated into \textit{non-lesion} (NL) and \textit{lesion}
(1--100\%). For the finer-grained three-class evaluation, they are
consolidated into the ordered classes \textit{normal} (NL),
\textit{mild--moderate} (1--50\%), and \textit{severe} (51--100\%).
At both granularities, the ground-truth label for each major branch is
determined by the highest severity among its sub-segments.
Panels~\textbf{A} and \textbf{B} of Figure~\ref{fig:report_metric_example}
show the resulting two- and three-class labels for one held-out test
case, respectively.

The report-generation prompt requires a structured JSON report with one
free-text field for each major branch. Within each field, findings are
reported at the segment level using numerical stenosis percentages or
ranges rather than qualitative severity terms alone. Vessels with no
diseased segments are reported as \texttt{normal}, whereas anatomically
absent or unvisualized vessels are reported as \texttt{nan}. Panel~\textbf{C}
of Figure~\ref{fig:report_metric_example} shows a CARDEA-generated report
produced under these instructions.

For each test case, only the generated free-text report is presented to
MedGemma-27B-IT~\cite{sellergren2025medgemma}, which extracts a
branch-level severity label for each major branch. A rule-based routine
then compares the extracted labels with the corresponding ground-truth
labels, pools the counts across all major branches and test cases, and
computes the macro-averaged $F_1$:
\begin{equation}
\text{VS-}F_1 = \frac{1}{|C|}\sum_{c \in C}
2\frac{P_cR_c}{P_c + R_c},
\end{equation}
where $C$ is the set of severity classes at the selected granularity,
and $P_c$ and $R_c$ are the conventional precision and recall for class
$c$. Because VS-$F_1$ is macro-averaged, each severity class contributes
equally, preventing the dominant normal class from overwhelming the
aggregate score.

To assess the stability of MedGemma label extraction under stochastic
decoding, we repeat the extraction five times at a sampling temperature
of 0.5 using the same fixed set of generated reports. The standard
deviation across runs is below 0.003 for every VS-$F_1$ entry reported
in the paper. A complementary consistency check on the full cohort
($n{=}412$) shows that MedGemma correctly extracts the label
\textit{normal} for every major branch from the always-normal reports,
yielding VS-$F_1$ scores identical to those obtained by directly
comparing fixed normal labels with the ground truth
(Table~\ref{tab:stage_wise_evolution}). Panel~\textbf{D} of
Figure~\ref{fig:report_metric_example} shows the fixed always-normal report
used for this check.

Alongside VS-$F_1$, we report ROUGE-L~\cite{lin2004rouge} and
BERTScore~\cite{zhang2019bertscore} as supplementary reference-based
text-similarity metrics. Because AngioCAD provides no reference reports,
we construct a reference for each test case by grouping the segment-level
labels under their corresponding major-branch fields without aggregating
them into branch-level labels (Figure~\ref{fig:report_metric_example}\textbf{E}).
The fixed always-normal report stress-tests these metrics' sensitivity
to wording overlap and normal-segment prevalence
(Section~\ref{sec:emergence}).

\begin{figure}[htbp]
\centering
\setlength{\fboxrule}{0.4pt}
\setlength{\fboxsep}{7pt}
\begin{minipage}[t]{0.48\linewidth}
\centering
\fbox{\begin{minipage}{0.90\linewidth}
\setstretch{1}
\textbf{A. Two-class VS-$F_1$ ground truth}\par
\vspace{0.4ex}\hrule\vspace{0.7ex}
{\small
\renewcommand{\arraystretch}{1.12}
\begin{tabularx}{\linewidth}{@{}>{\bfseries}lX@{}}
LM  & lesion \\
LAD & non-lesion \\
LCX & lesion \\
RCA & lesion
\end{tabularx}}
\end{minipage}}
\end{minipage}\hfill
\begin{minipage}[t]{0.48\linewidth}
\centering
\fbox{\begin{minipage}{0.90\linewidth}
\setstretch{1}
\textbf{B. Three-class VS-$F_1$ ground truth}\par
\vspace{0.4ex}\hrule\vspace{0.7ex}
{\small
\renewcommand{\arraystretch}{1.12}
\begin{tabularx}{\linewidth}{@{}>{\bfseries}lX@{}}
LM  & mild--moderate \\
LAD & normal \\
LCX & mild--moderate \\
RCA & severe
\end{tabularx}}
\end{minipage}}
\end{minipage}

\vspace{1.2ex}
\fbox{\begin{minipage}{0.95\linewidth}
\setstretch{1}
\textbf{C. CARDEA-generated report}\par
\vspace{0.4ex}\hrule\vspace{0.7ex}
{\scriptsize\ttfamily\raggedright
\{\par
\begin{tabularx}{\linewidth}{@{\quad}l@{\ }>{\raggedright\arraybackslash}X@{}}
"lm":  & "30-50\% stenosis", \\
"lad": & "normal", \\
"lcx": & "normal", \\
"rca": & "Severe 70-90\% stenosis in proximal and mid segments, mild 0-50\% in distal"
\end{tabularx}
\}\par}
\end{minipage}}

\vspace{1.2ex}
\fbox{\begin{minipage}{0.95\linewidth}
\setstretch{1}
\textbf{D. Fixed always-normal report}\par
\vspace{0.4ex}\hrule\vspace{0.7ex}
{\scriptsize\ttfamily\raggedright
\{\par
\begin{tabularx}{\linewidth}{@{\quad}l@{\ }>{\raggedright\arraybackslash}X@{}}
"lm":  & "normal/nan", \\
"lad": & "proximal normal/nan; mid normal/nan; distal normal/nan; 1st diagonal normal/nan; 2nd diagonal normal/nan", \\
"lcx": & "proximal normal/nan; mid normal/nan; distal normal/nan; om normal/nan", \\
"rca": & "proximal normal/nan; mid normal/nan; distal normal/nan; pda normal/nan; plb normal/nan"
\end{tabularx}
\}\par}
\end{minipage}}

\vspace{1.2ex}
\fbox{\begin{minipage}{0.95\linewidth}
\setstretch{1}
\textbf{E. ROUGE-L/BERTScore reference}\par
\vspace{0.4ex}\hrule\vspace{0.7ex}
{\scriptsize\ttfamily\raggedright
\{\par
\begin{tabularx}{\linewidth}{@{\quad}l@{\ }>{\raggedright\arraybackslash}X@{}}
"lm":  & "0-50\% stenosis", \\
"lad": & "proximal normal/nan; mid normal/nan; distal normal/nan; 1st diagonal normal/nan; 2nd diagonal normal/nan", \\
"lcx": & "proximal normal/nan; mid 0-50\% stenosis; distal normal/nan; om normal/nan", \\
"rca": & "proximal 0-50\% stenosis; mid 50-100\% stenosis; distal 0-50\% stenosis; pda normal/nan; plb 0-50\% stenosis"
\end{tabularx}
\}\par}
\end{minipage}}
\caption{Example report-evaluation representations for one held-out AngioCAD test case. \textbf{A,} Branch-level ground-truth labels for the primary two-class VS-$F_1$, aggregated from AngioCAD's original multilevel stenosis grades assigned to individual coronary segments; \textbf{B,} corresponding labels for the finer-grained three-class VS-$F_1$; \textbf{C,} report generated by CARDEA during inference on the same case; \textbf{D,} fixed always-normal report used for the consistency check and baseline evaluation; and \textbf{E,} reference constructed for ROUGE-L and BERTScore by grouping AngioCAD's segment-level stenosis annotations under their corresponding major branches without aggregating them into branch-level labels.}
\label{fig:report_metric_example}
\end{figure}

\subsection{Training Configuration}
\label{sec:appendix_training}

We use Qwen3-VL-30B-A3B-Thinking as the backbone and conduct
training on a single node equipped with eight NVIDIA~B200 GPUs.
The Align and Cold Start stages are implemented with
LlamaFactory~\cite{zheng2024llamafactory}, whereas RLVR is implemented
with EasyR1~\cite{zheng2025easyr1,sheng2024hybridflow}.
Table~\ref{tab:training_config} reports the principal hyperparameters
and distributed configuration for each stage.

\begin{table}[htbp]
\centering
\caption{Stage-specific training configuration.}
\label{tab:training_config}
\small
\renewcommand{\arraystretch}{1.15}
\begin{tabularx}{\linewidth}{@{}p{0.29\linewidth}>{\raggedright\arraybackslash}X@{}}
\toprule
\textbf{Setting} & \textbf{Value} \\
\midrule
\multicolumn{2}{@{}l}{\textbf{SFT: Align and Cold Start}} \\
Adaptation & LoRA (rank 32, $\alpha=64$, dropout 0) \\
Optimizer & AdamW \\
Learning rate & $1\times10^{-4}$ for Align; $5\times10^{-5}$ for Cold Start \\
Global batch size & 64 \\
Training duration & 2 epochs per stage \\
Distributed configuration & DeepSpeed ZeRO Stage 2 \\
\midrule
\multicolumn{2}{@{}l}{\textbf{RLVR}} \\
Algorithm & SAPO \\
Optimizer & AdamW \\
Learning rate & $1\times10^{-6}$ \\
Rollout batch size & 32 \\
Responses per prompt & 13 \\
Rollout temperature & 0.8 \\
Distributed configuration & FSDP for policy updates \\
\bottomrule
\end{tabularx}

\vspace{0.5ex}
{\footnotesize\setstretch{1}\raggedright
SFT, supervised fine-tuning; LoRA, low-rank adaptation~\cite{hu2021lora};
AdamW, Adam with decoupled weight decay~\cite{loshchilov2017decoupled};
RLVR, reinforcement learning with verifiable rewards; SAPO, Soft Adaptive
Policy Optimization~\cite{gao2025soft}; ZeRO, Zero Redundancy
Optimizer~\cite{rajbhandari2020zero}; FSDP, Fully Sharded Data
Parallel~\cite{zhao2023pytorch}.\par}
\end{table}

\subsection{Reward Functions}
\label{sec:appendix_rewards}

\subsubsection{Total Reward}
\label{sec:appendix_reward_total}

During RLVR, we assign each rollout a format reward, a task-specific
accuracy reward, and a CoB reward. We denote these terms by
$R_{\mathrm{fmt}}$, $R_{\mathrm{acc}}$, and $R_{\mathrm{CoB}}$.
Their weights are $w_{\mathrm{fmt}}$, $w_{\mathrm{acc}}$, and
$w_{\mathrm{CoB}}$. The total reward is
\begin{equation}
R_{\mathrm{total}} = w_{\mathrm{fmt}} R_{\mathrm{fmt}} + \mathbbm{1}(R_{\mathrm{fmt}}>0)\big[\, w_{\mathrm{acc}} R_{\mathrm{acc}} + w_{\mathrm{CoB}} R_{\mathrm{CoB}} \,\big].
\end{equation}
In our configuration, the format weight is 0.2 and the accuracy weight
is 0.8. The CoB weight is 1.2 for study-level tasks and 0 for
single-view tasks. The format reward also gates the other components.
If it is zero, neither accuracy nor CoB contributes to the total reward.

\subsubsection{Format Reward}
\label{sec:appendix_reward_fmt}

The format reward is granted when the response follows the required
output format:
\begin{equation}
R_{\mathrm{fmt}} = \mathbbm{1}\big(\text{reasoning and answer are both present and separately parseable}\big).
\end{equation}

\subsubsection{Task-Specific Accuracy Reward}
\label{sec:appendix_reward_acc}

The accuracy reward is defined separately for classification, detection,
and keyframe selection.

For classification, let $\hat{y}$ and $y_{gt}$ denote the
predicted and ground-truth classes, respectively. The reward is binary
accuracy per sample:
\begin{equation}
R_{\mathrm{acc}}^{cls} = \mathbbm{1}(\hat{y} = y_{gt}).
\end{equation}

For detection, the reward reuses the one-to-one matching procedure of
Section~\ref{sec:appendix_f1iou} but replaces the binary match outcome
with a graded score. A roughly correct box therefore receives partial
credit. For each predicted box $b$, let $g$ denote its matched
ground-truth box when a match exists. The localization score $s(b)$ is
\begin{equation}
s(b) =
\begin{cases}
0, & b \text{ unmatched, or } \operatorname{IoU}(b_{box}, g_{box}) \le 0.2,\\[3pt]
\dfrac{\operatorname{IoU}(b_{box}, g_{box}) - 0.2}{0.7 - 0.2},
& 0.2 < \operatorname{IoU}(b_{box}, g_{box}) < 0.7,\\[5pt]
1, & \operatorname{IoU}(b_{box}, g_{box}) \ge 0.7
\end{cases}.
\end{equation}
Summing these scores gives fractional counts:
\begin{equation}
TP = \sum_{b \in B_{pred}} s(b), \quad
FP = |B_{pred}| - TP, \quad
FN = |B_{gt}| - TP.
\end{equation}
These counts define precision and recall for every rollout:
\begin{equation}
P =
\begin{cases}
\dfrac{TP}{TP + FP}, & TP + FP > 0,\\[4pt]
0, & TP + FP = 0,
\end{cases}
\qquad
R =
\begin{cases}
\dfrac{TP}{TP + FN}, & TP + FN > 0,\\[4pt]
0, & TP + FN = 0.
\end{cases}
\end{equation}
The detection accuracy reward is
\begin{equation}
R_{\mathrm{acc}}^{det} =
\begin{cases}
1, & B_{pred} = B_{gt} = \emptyset,\\[3pt]
2 \cdot \dfrac{P \cdot R}{P + R}, & P + R > 0,\\[5pt]
0, & \text{otherwise}
\end{cases}.
\end{equation}
The detection scores
in Table~\ref{tab:foundational_tasks} use the hard
$F_1$@IoU$\ge$0.5 of Section~\ref{sec:appendix_f1iou} and pool
boxes over the complete test set. The graded reward above is used only
during training and is computed separately for each rollout.

For keyframe selection, the prompt asks the model to return a JSON
object containing a list of diagnostically usable frame indices and one
optimal-frame index. We denote the predicted and ground-truth
usable-frame sets by $S_{pred}$ and $S_{gt}$, and the predicted
optimal-frame index by $idx_{best}^{pred}$. The precision and recall of
the predicted usable-frame set are
\begin{equation}
P = \frac{|S_{pred} \cap S_{gt}|}{|S_{pred}|}, \qquad
R = \frac{|S_{pred} \cap S_{gt}|}{|S_{gt}|}.
\end{equation}
The keyframe-selection accuracy reward adds an optimal-frame bonus to
their harmonic mean:
\begin{equation}
R_{\mathrm{acc}}^{key} = 2 \frac{P R}{P + R}
+ \mathbbm{1}(idx_{best}^{pred} \in S_{pred} \cap S_{gt}).
\end{equation}
The bonus is awarded when the predicted optimal frame belongs to both
usable-frame sets.

\subsubsection{Chain-of-Box Reward}
\label{sec:appendix_reward_cob}

The CoB reward encourages the model to include spatial evidence in
its reasoning trace. For a reasoning trace $\tau$, let $g(\tau)$
indicate whether at least one valid bounding box can be parsed from it:
\begin{equation}
g(\tau) = \mathbbm{1}\big(\text{at least one valid bounding box can be parsed from } \tau\big).
\end{equation}
The CoB reward is gated on answer correctness through the classification
reward defined in Section~\ref{sec:appendix_reward_acc}:
\begin{equation}
R_{\mathrm{CoB}} = g(\tau)\cdot R_{\mathrm{acc}}^{cls}.
\end{equation}
The unconditional ablation of Section~\ref{sec:ablations} removes
the gate and rewards the box by itself:
\begin{equation}
R_{\mathrm{CoB}}^{\mathrm{uncond}} = g(\tau).
\end{equation}

\FloatBarrier
\section{Supplementary Results}
\label{sec:appendix_results}

\subsection{Repeated-Sampling CoB Coverage}
\label{sec:appendix_cob_coverage}

\begin{figure}[htbp]
\centering
\begin{minipage}{0.48\textwidth}
  \centering
  \includegraphics[width=\linewidth]{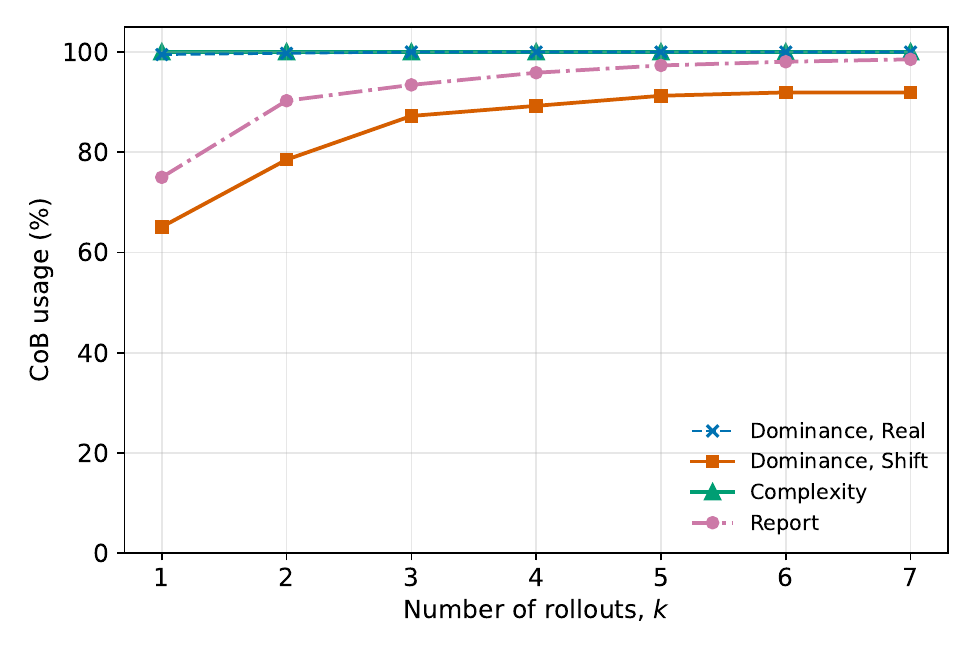}
  \vspace{0.5ex}
  \centerline{\textbf{A}}
\end{minipage}
\hfill
\begin{minipage}{0.48\textwidth}
  \centering
  \includegraphics[width=\linewidth]{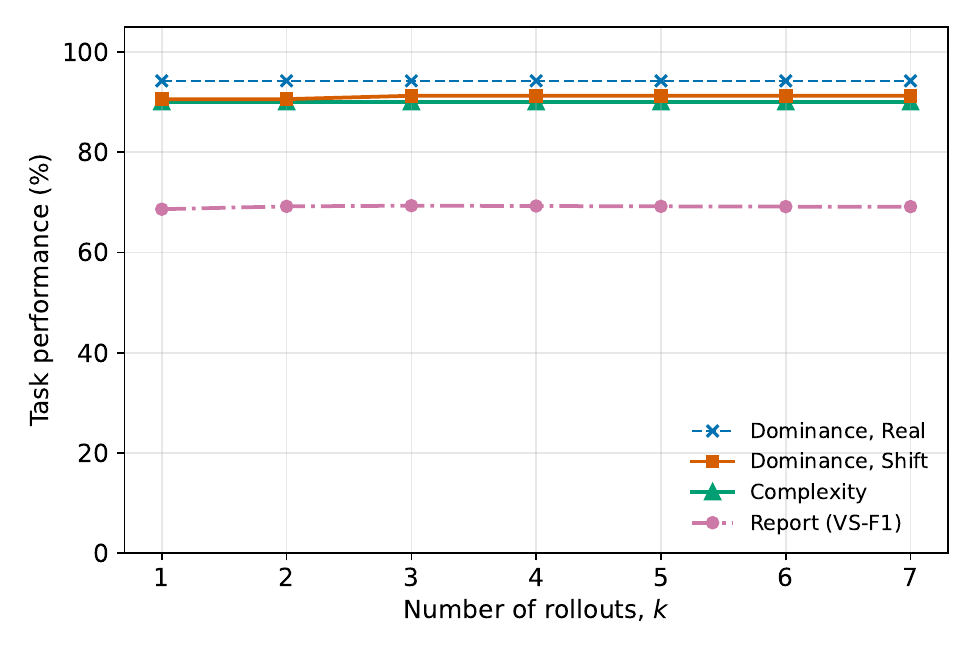}
  \vspace{0.5ex}
  \centerline{\textbf{B}}
\end{minipage}
\caption{Repeated-Sampling CoB Coverage and Task Performance. For each study, the first rollout with a valid in-trace bounding box among the first $k$ was selected; if none had one, rollout 1 was retained. \textbf{A,} CoB usage@$k$, the percentage of selected rollouts containing a box. \textbf{B,} Task performance@$k$, measured by exact-match accuracy for the classification tasks and two-class VS-$F_1$ for report generation. Curves show CARDEA with thinking enabled at sampling temperature 0.5 on the official CoronaryDominance Real ($n{=}400$) and Domain Shift ($n{=}149$) test sets, the CardioSyntax Complexity test set ($n{=}60$), and zero-shot report generation on the full AngioCAD cohort ($n{=}412$); $k{=}1$ matches the corresponding single-pass results in the main text.}
\label{fig:appendix_cob_coverage}
\end{figure}

The incomplete CoB coverage noted in the Discussion is a property of a single pass and is partly recoverable by sampling. Exploiting the model's decoding stochasticity (thinking enabled, sampling temperature 0.5), we drew seven independent rollouts per study for the three trained-task test sets and for zero-shot report generation on the full AngioCAD cohort ($n{=}412$). Among the first $k$, we select the first rollout whose reasoning trace contains a valid CoB box, and keep the first rollout if none of the first $k$ does. CoB usage@$k$ is the fraction of selected rollouts containing a box. Task performance@$k$ is exact-match accuracy for the classification tasks and two-class VS-$F_1$ for report generation; at $k{=}1$, both measures reproduce the corresponding single-pass values reported in the main text. As Figure~\ref{fig:appendix_cob_coverage} shows, Real Distribution usage rose from 99.5\% to 100.0\% with accuracy unchanged at 94.3\%, whereas Complexity remained at 100.0\% usage and 90.0\% accuracy. By contrast, Domain Shift usage rose from 65.1\% at $k{=}1$ to 92.0\% at $k{=}7$, while accuracy changed only from 90.6\% to 91.3\%. For report generation, CoB usage increased from 75.0\% at $k{=}1$ to 98.5\% at $k{=}7$, while two-class VS-$F_1$ changed only from 68.6\% to 69.1\%. The small fluctuations should not be interpreted as evidence of improved or degraded report quality because they may reflect both rollout-sampling and judge variability. Thus, repeated sampling improved CoB coverage on the lower-coverage evaluations without materially changing task performance.

\FloatBarrier
\section{Extended Limitations}
\label{sec:appendix_extended_limitations}

The Discussion summarizes the key limitations of the study. A complete account follows.

\subsection{Training and Data Limitations}

Public training datasets such as ARCADE and CADICA mainly annotate the location and severity of coronary stenoses. However, comprehensive diagnostic interpretation also requires characterization of additional lesion features (e.g., calcification, thrombus, dissection, and bifurcation involvement). These features may affect lesion assessment and subsequent interventional planning. Recognizing procedural equipment (e.g., catheters and guidewires) and implanted devices (e.g., coronary stents) may also be important for understanding the context in which images were acquired. For example, the DeepCORO-CLIP pipeline~\cite{harrabi2026deepcoroclip} identifies non-coronary structures and stent presence. It also uses the first appearance of interventional equipment to retain only diagnostic videos. Such information could help an operational system select diagnostically relevant content from raw procedural recordings during deployment. CARDEA and its current evaluation may therefore not fully capture the range of angiographic findings and procedural contexts encountered in routine coronary angiography.

The CoronaryDominance and CardioSyntax training sets were constructed with a preliminary LVLM that selected keyframes and filtered views (Appendix~\ref{sec:appendix_datasets}). Each study was limited to 10 static keyframes, so the training data support cross-view reasoning but not the temporal dynamics of contrast flow. Any systematic preference for particular projection angles or contrast densities may therefore be reflected in those sets, but we did not measure its downstream effect.

The AngioCAD RCA stenosis task labels any visible narrowing ($>0\%$), whereas ARCADE uses a $\ge 50\%$ threshold. CADICA supplied sub-50\% lesion labels during feature alignment, but we did not test whether this prior exposure affected performance on AngioCAD.

CARDEA's training used one run per stage, so the bootstrap CIs capture test-set sampling variability but not variation across training runs.

\subsection{Evaluation Limitations}

The evaluation of trained closed-ended tasks is limited by small test sets. Keyframe selection includes 48 videos from 5 patients, while complexity assessment includes 60 studies. Both may be underpowered and should be considered preliminary. Direct comparability is also limited by two output conversions. We converted DeepCoro's Algorithm~4 segmentation masks to bounding boxes, possibly understating its detection score. For complexity assessment, we binarized the cardiologists' continuous SYNTAX scores rather than using predictions produced natively for the binary task. Although the CIs overlap, CARDEA's Macro $F_1$ point estimate is lower than either cardiologist's, possibly reflecting reduced sensitivity to the minority high-risk class, where missed high-risk studies are the main clinical concern. Statistically non-distinguishable results do not establish equivalence.

The valid-views subset retains only AngioCAD studies in which CARDEA identified both an LCA and an RCA view. Although this provides potentially more complete view coverage, it may introduce selection bias by excluding more difficult studies. Results on the complete cohort are therefore primary.

The report-generation evaluation uses MedGemma-27B-IT to extract per-vessel severity labels from each generated report. The ground-truth labels and final $F_1$ are derived deterministically, so only label extraction depends on MedGemma (Appendix~\ref{sec:appendix_eval_vsf1}). The parser exactly reproduces the naive floor on always-normal inputs, but its agreement with human annotators on ambiguous free-text reports was not measured.

VS-$F_1$ also coarsely measures a single dimension of report quality. It scores branch-level diagnostic agreement after the 15 annotated segments are collapsed into four branches by maximum severity, so a report that assigns a lesion to the wrong segment within a branch, or omits a second lesion in the same branch, is still scored as correct. Expression, including fluency, readability, and terminological precision, enters none of our metrics. The report metric also lacks a human-expert reference. Report scores should therefore be interpreted relative to the naive floor and across stages rather than as absolute measures of report quality.

\subsection{Model and Reasoning Limitations}

CoB coverage was lower on the Domain Shift test set and on zero-shot report generation than on the in-distribution trained tasks. Repeated sampling partly recovered coverage without materially changing task performance (Appendix~\ref{sec:appendix_cob_coverage}). Before deployment, CoB coverage should therefore be evaluated on data representative of the intended setting.

Box presence does not establish box faithfulness. Our evaluation measures box frequency and area and relates box presence to final-answer correctness, but it does not test whether a generated box accurately localizes the structure it names. The boxes expose spatial claims for human inspection but are not verified annotations. Their faithfulness requires clinically supervised assessment.

Near-universal coverage may require rejection sampling to select correct-answer trajectories with valid CoBs, followed by further SFT~\cite{deepseek2025r1}. The same offline route could also serve open-ended tasks, filtering repeated CARDEA rollouts by complementary measures of report-quality dimensions that VS-$F_1$ leaves unscored, or by expert review, and fine-tuning on the survivors. Because this filtering is performed offline rather than used as an optimization reward, it avoids directly optimizing the policy against an open-ended reward model, the setting whose hacking risk motivated excluding open-ended tasks from RLVR.

\subsection{Clinical and Operational Readiness}

Served with vLLM~\cite{kwon2023efficient} on four NVIDIA~B200 GPUs with tensor parallelism, end-to-end processing from raw multi-view videos through report generation took a median of roughly 6 seconds per study across 30 randomly sampled AngioCAD studies. In the first pass, component tasks run in parallel, so its wall-clock time is determined by the slowest task. These values are hardware-dependent. On less capable hardware, autoregressive decoding of a full CoB reasoning trace may be substantially slower than a single discriminative forward pass. CARDEA is therefore unsuitable for settings that require sub-second responses, including intra-procedural real-time guidance. Latency should be evaluated on the intended hardware before deployment.

CARDEA has not been validated in a clinical setting. Prospective evaluation against domain experts is a prerequisite for clinical use and should include blinded comparison with physician-authored reports.

\FloatBarrier
\section{Qualitative Inference Examples}
\label{sec:appendix_figures}

For each task, the example shown was drawn at random from the corresponding test set: the CoronaryDominance Real Distribution subset for dominance classification, the CardioSyntax test set for complexity assessment, and the held-out AngioCAD cohort for report generation. Sampling was restricted to cases whose reasoning trace contained at least one bounding box; for the two classification tasks, it was further restricted to cases in which CARDEA produced the correct final answer.

\clearpage
\begin{figure}[p]
\centering
\includegraphics[height=0.86\textheight,width=\textwidth,keepaspectratio]{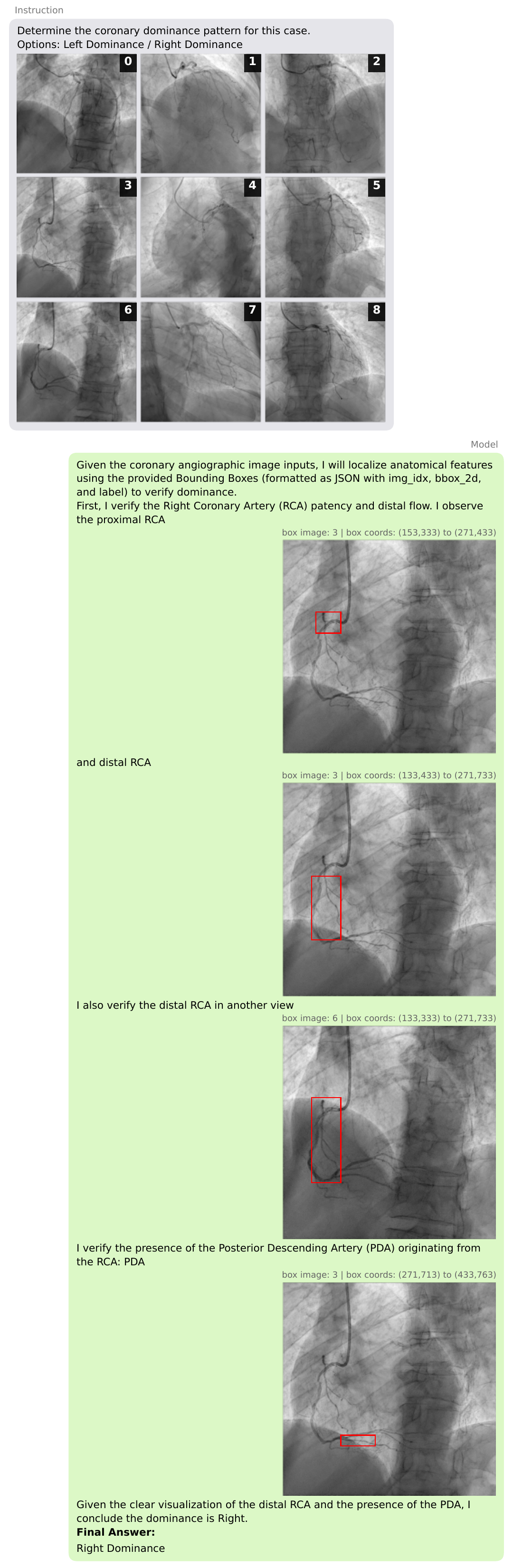}
\caption{Qualitative CARDEA inference example for coronary dominance classification. The example shows the multi-view input, the Chain-of-Box (CoB) reasoning trace with spatially grounded coronary structures, and the final dominance prediction.}
\label{fig:infer_dominant}
\end{figure}

\clearpage
\begin{figure}[p]
\centering
\includegraphics[height=0.86\textheight,width=\textwidth,keepaspectratio]{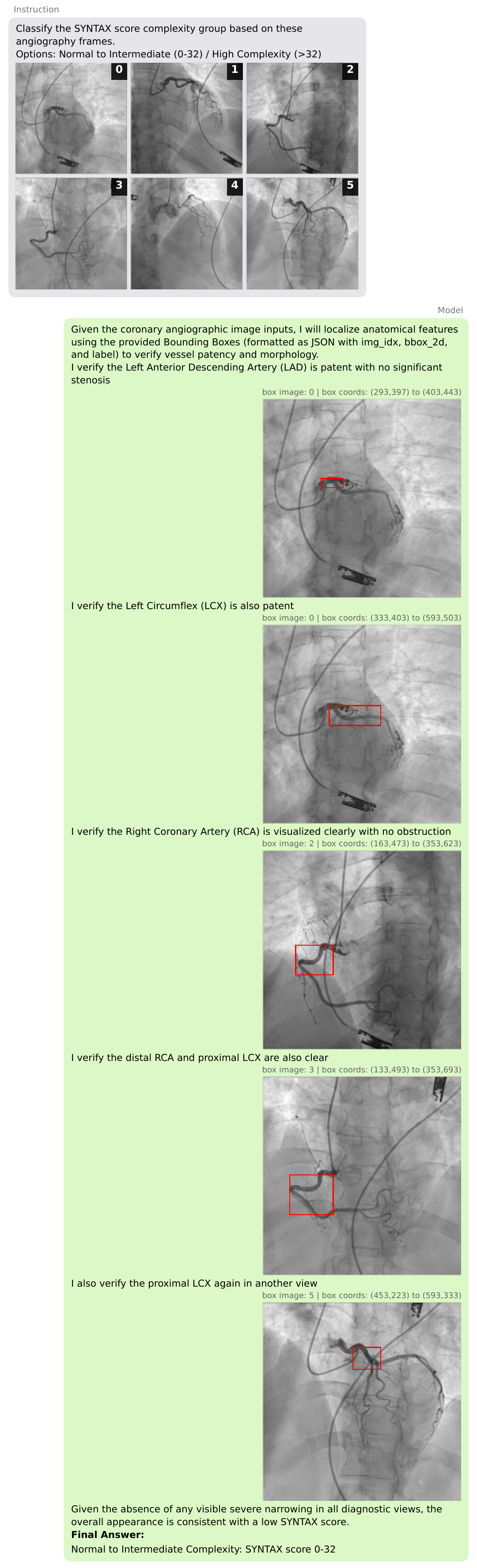}
\caption{Qualitative CARDEA inference example for anatomical complexity assessment. The example shows the multi-view input, the Chain-of-Box (CoB) reasoning trace with spatial grounding of coronary vessels and lesions, and the final complexity category.}
\label{fig:infer_complexity}
\end{figure}

\clearpage
\begin{figure}[p]
\centering
\includegraphics[height=0.86\textheight,width=\textwidth,keepaspectratio]{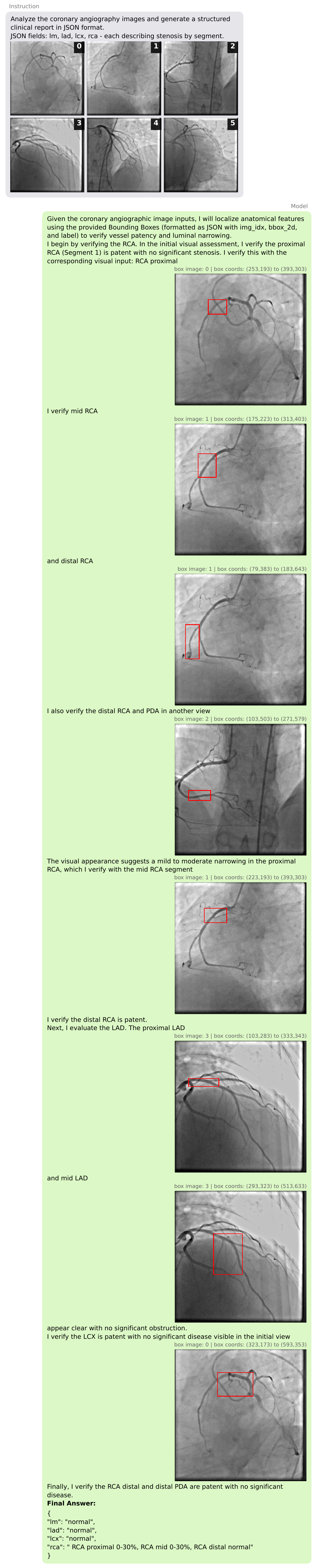}
\caption{Qualitative CARDEA inference example for structured report generation. The example shows the multi-view input, the Chain-of-Box (CoB) reasoning trace with vessel-level spatial grounding, and the resulting branch-level JSON report.}
\label{fig:infer_report}
\end{figure}

\end{document}